\documentclass[10pt,journal,compsoc]{IEEEtran}
\ifCLASSOPTIONcompsoc
  \usepackage[nocompress]{cite}
\else
  \usepackage{cite}
\fi

\usepackage[pdftex]{graphicx}
\usepackage{makecell}
\usepackage{booktabs}
\usepackage{multirow}
\usepackage{amsfonts}
\usepackage{amsmath}
\usepackage{mathrsfs}
\usepackage{bm}
\usepackage{cases}
\usepackage{balance}
\usepackage{algorithm}
\usepackage{algorithmic}

\newtheorem{myDef}{Definition}
\usepackage{enumerate}

\begin{document}
\title{Link Prediction on N-ary Relational Data\\ Based on Relatedness Evaluation}
\author{Saiping~Guan,~Xiaolong~Jin,~Jiafeng~Guo,~Yuanzhuo~Wang, and~Xueqi~Cheng
\IEEEcompsocitemizethanks{\IEEEcompsocthanksitem 
The authors are with the School of Computer Science and Technology, University of Chinese Academy of Sciences, Beijing 100049, China, and also with the CAS Key Lab of Network Data Science and Technology, Institute of Computing Technology, Chinese Academy of Sciences, Beijing 100864, China. E-mail: \{guansaiping,jinxiaolong,guojiafeng,wangyuanzhuo,cxq\} @ict.ac.cn.
}
\thanks{Manuscript received 19 May 2020; revised 19 Mar. 2021; accepted 7 Apr. 2021. (Corresponding author: Saiping Guan.)}}

\markboth{Journal of \LaTeX\ Class Files,~Vol.~14, No.~8, August~2015}%
{Shell \MakeLowercase{\textit{Guan et al.}}: Link Prediction on N-ary Relational Data Based on Relatedness Evaluation}

\IEEEtitleabstractindextext{%
\begin{abstract}
With the overwhelming popularity of Knowledge Graphs (KGs), researchers have poured attention to link prediction to fill in missing facts for a long time. However, they mainly focus on link prediction on binary relational data, where facts are usually represented as triples in the form of (head entity, relation, tail entity). In practice, n-ary relational facts are also ubiquitous. When encountering such facts, existing studies usually decompose them into triples by introducing a multitude of auxiliary virtual entities and additional triples. These conversions result in the complexity of carrying out link prediction on n-ary relational data. It has even proven that they may cause loss of structure information. To overcome these problems, in this paper, we represent each n-ary relational fact as a set of its role and role-value pairs. We then propose a method called NaLP to conduct link prediction on n-ary relational data, which explicitly models the relatedness of all the role and role-value pairs in an n-ary relational fact. We further extend NaLP by introducing type constraints of roles and role-values without any external type-specific supervision, and proposing a more reasonable negative sampling mechanism. Experimental results validate the effectiveness and merits of the proposed methods.
\end{abstract}

}

\maketitle
\IEEEdisplaynontitleabstractindextext
%
\IEEEpeerreviewmaketitle

\IEEEraisesectionheading{\section{Introduction}\label{sec:introduction}}
\IEEEPARstart{S}{ince} Google announced Knowledge Graph (KG) in 2012, KG has been increasingly popular in these years. And link prediction on KG has caught much attention for a long time, to complete KG and further promote KG based applications. Usually, a KG is represented as a set of triples in the form of (head entity, relation, tail entity), where n-ary relational facts are decomposed into multiple triples via introducing virtual entities, such as the Compound Value Type (CVT) entities in Freebase~\cite{Freebase}. Actually, n-ary relational facts are not in the minority. As an example, in Freebase, more than $1/3$ of its entities are involved in n-ary relational facts~\cite{m-TransH}. Based on the public KGs of the above binary form, researchers have developed numerous methods for link prediction~\cite{nickel2016review,wang2017review,GE_survey}, including the tasks of head entity prediction, relation prediction, and tail entity prediction.

However, manipulating n-ary relational facts into triples has some deficiencies. Firstly, it makes link prediction on n-ary relational data need to consider many triples. Since existing link prediction methods on triples, like the above methods, usually model triples one by one individually and correspondingly conduct link prediction on only a single triple each time, it is intricate to carry out link prediction involving more than one triple. Secondly, there is a loss of structure information in some of the conversions from n-ary relational facts to triples~\cite{m-TransH}, which may lead to inaccurate prediction. Thirdly, the added virtual entities, along with the corresponding triples, bring in more parameters to be learned, and there is a data sparsity problem.

With these considerations in mind, in this paper, each n-ary relational fact is represented as a set of its role and role-value pairs, denoted as r:v pairs for short, instead of being converted into multiple triples. Hence, no additional entities and triples are introduced, and all the structure information is retained. Then, we propose a method to handle link prediction on n-ary relational data. Specifically, link prediction on n-ary relational data is to predict the missing role/role-value of an n-ary relational fact.

Recently, there are also a few studies for link prediction on n-ary relational data. In m-TransH~\cite{m-TransH}, an n-ary relation is defined by the mappings from a sequence of roles to their role-values. Each specific mapping is an n-ary relational fact. Then, a translation based method is proposed to model these facts. RAE~\cite{RAE} improves m-TransH with the modeling of the likelihood that two role-values co-participate in a common n-ary relational fact. Although RAE achieves favorable performance, it does not consider the roles explicitly when evaluating the above likelihood. Actually, under different sequences of roles (corresponding to different relations), the relatedness of two role-values is greatly different. For example, under the role sequence $(person, award, point\ in\ time, together\ with)$ (in this paper, the examples are all derived from Wikidata\footnote{https://www.wikidata.org/wiki/Wikidata:Main\_Page\label{footnote:wikidata}}), $Marie$ $Curie$ and $Henri\ Becquerel$ are more related than under the role sequence $(person, spouse, start\ time, end\ time,$ $place\ of\ marriage)$, since they won $Nobel\ Prize\ in$ $Physics$ in $\textit{1903}$ together. To address these problems, our method explicitly models the relatedness of the r:v pairs involved in an n-ary relational fact. The above example also elucidates its necessity. Some newly proposed methods~\cite{HypE,GETD,HINGE,NeuInfer} further apply position-specific convolution~\cite{HypE}, tensor decomposition techniques~\cite{GETD}, or structure-aware methods~\cite{HINGE,NeuInfer}. They do not consider any type constraint.

Since for a valid n-ary relational fact, the expected type of the role-value in the view of the role should be compatible with the actual type of the role-value in each r:v pair, we combine these signals without explicit external supervision from a type catalog. Thus, NaLP is extended to tNaLP (type-NaLP). Furthermore, we equip NaLP and tNaLP with a more reasonable negative sampling mechanism and thus extend them to NaLP\textsuperscript{+} and tNaLP\textsuperscript{+}, respectively. 

In general, the main contributions of this paper are:
\begin{itemize}
	\item We advocate a representation form for n-ary relational facts, which represents each n-ary relational fact as a set of its r:v pairs.
	\item We propose a link prediction method NaLP on n-ary relational data, which captures the relatedness of all the r:v pairs in an n-ary relational fact explicitly.
	\item We further introduce type constraints of roles and role-values in an unsupervised manner, and propose a more reasonable negative sampling mechanism. Thus, NaLP is extended to tNaLP, NaLP\textsuperscript{+}, and then tNaLP\textsuperscript{+}, respectively.
	\item Experimental results and analyses testify the effectiveness and superiority of the proposed methods.
\end{itemize}

In addition, as publicly available n-ary relational datasets are limited, we build a practical one, WikiPeople, based on Wikidata~\cite{Wikidata}. We further derive some new ones, WikiPeople-n, WikiPeople-0bi, WikiPeople-50bi, and WikiPeople-100bi, from WikiPeople with different percentages of binary relational facts. They have been available on github\footnote{https://github.com/gsp2014/WikiPeople} for further research. We have also opened our source codes of the proposed methods\footnote{https://github.com/gsp2014/NaLP}.

\section{Related Works}
\label{sec:related_works}
\subsection{Link Prediction on Binary Relational Data}
The rapid development of KG has triggered the spring up of a great many methods, including tensor/matrix based methods, translation based ones, and neural network based ones, for link prediction on binary relational data.

Among tensor/matrix based methods~\cite{RESCAL,nickel2014reducing,krompabeta2015type,ComplEx,RSTE,ComplEx_modify,ComplEx_modify_cn,TypeComplex}, RESCAL~\cite{RESCAL} views a KG as a three-way tensor, where each way corresponds to head entities, relations, or tail entities. The entry corresponds to the score indicating the validity of the triple formed by the thee ways. Through minimizing the reconstruction error of the tensor, the embeddings of entities and relations are learned. Then, the reconstructed tensor from the learned embeddings is used to conduct link prediction, with triples corresponding to entries of high scores as valid ones. Based on RESCAL, type constraints are further integrated in~\cite{krompabeta2015type} via indexing only those embeddings of entities for each relation that agree with specific type constraint. Similar to RESCAL, ComplEx~\cite{ComplEx} constructs a matrix for each relation, which is factorized and optimized via minimizing the reconstruction error to learn the embeddings. Differently, complex values are used to define the embeddings, so as to deal with antisymmetric relations effectively. Upon ComplEx, TypeComplex~\cite{TypeComplex} enhances each base factorization with two type compatibility terms between entity-relation pairs (one for the head entity and the other for the tail entity).  

Translation based methods are derived from TransE~\cite{TransE}. Its translation assumption advocates that valid triples can be viewed as relational translations from head entities to tail entities. Based on this assumption, different types of translations and accordingly different score functions measuring the similarity between the relational translation results of the head entities and the target tail entities are defined~\cite{TransE,TransH,TransR,TransD,PTransE,TransG,TranSparse,TransA,TKRL,TransT,SSE,SSE_extend,TorusE,TorusE_extend,TAPR}. The embeddings of entities and relations are then learned via minimizing score function based loss, which encourages valid triples to have much larger scores than invalid ones. Among these methods, TransE, performs translations in a shared space of entities and relations. TransH~\cite{TransH} further thinks translations should be conducted in relation-specific hyperplanes and relates each relation with a hyperplane. Then, entities are projected into the hyperplanes of relations before translations. TKRL~\cite{TKRL} takes advantage of hierarchical entity types in the construction of projection matrices for entities. And in negative sampling, it tends to select entities having the same types; In evaluation, it removes the candidates which do not follow type constraints. TransT~\cite{TransT} constructs relation types from entity types and utilizes type-based semantic similarity of the related entities and relations to capture prior distributions of entities and relations. With these prior distributions, it generates semantic vectors for entities in different contexts, which are learned following TransE. Then, it estimates the posterior probability of entity and relation predictions. TAPR~\cite{TAPR} uses types to enrich the embeddings of entities and relations, and describes a type-level attention to select the most relevant type of the given entity in a specific triple.

Inspired by the excellent performance of neural networks in various applications, they have been introduced into link prediction on binary relational data and have achieved extremely promising results~\cite{ER-MLP,ProjE,R-GCN,ConvE,SENN,ConvKB,SACN,C-RNN,CompGCN}. Among them, the R-GCN~\cite{R-GCN} method introduces relational Graph Convolutional Networks (GCN) as its encoder to get the embeddings of entities, and uses the existing method~\cite{DistMult} as its decoder to compute the scores of triples with the embeddings of entities from the encoder and the embeddings of relations randomly initialized. CompGCN~\cite{CompGCN} further learns the embeddings of relations in the GCN encoder. Differently, SENN~\cite{SENN} integrates the prediction tasks of head entities, relations, and tail entities in a unified Fully Connected Network (FCN) framework based on shared embeddings. ConvKB~\cite{ConvKB} represents each triple as a three-column matrix. This matrix is fed into a convolution layer to get feature maps, which are then concatenated into a single feature vector representing the triple. After passing it to a fully connected layer, ConvKB obtains a score indicating the validity of the triple.  

\subsection{Link Prediction on N-ary Relational Data}
As n-ary relational facts are complicated and usually tricky to deal with, most of existing studies convert them into triples. Studies for link prediction on n-ary relational data directly are relatively much scarcer and newer. 

The m-TransH~\cite{m-TransH} method generalizes TransH~\cite{TransH} on binary relational data to n-ary relational data. It presents a mathematical definition of n-ary relations as the mappings from sequences of roles to their role-values. Each n-ary relational fact is a specific mapping of the corresponding relation. Then, the score function of an n-ary relational fact is defined by the weighted sum of the projection results from its role-values to its relation hyperplane, where the weights are the real numbers projected from its roles. Based on m-TransH, RAE~\cite{RAE} introduces the relatedness of role-values, i.e., the likelihood that two role-values co-participate in a common n-ary relational fact, and adds this relatedness loss with a weight hyper-parameter to the embedding loss of m-TransH. Although with the additional modeling of the relatedness of role-values, RAE outperforms m-TransH, it does not look into the roles when computing the relatedness. Since roles are also a fundamental part of n-ary relational facts, taking them into consideration may make a difference as illustrated afore. To better capture the role that the entity plays in a given relation, HypE~\cite{HypE} further convolves each entity appearing at the specific position in a given tuple with the set of position-specific filters. GETD~\cite{GETD} is the first tensor based method for n-ary relational link prediction. However, GETD only focuses on handling n-ary relational facts of the same arities. HINGE~\cite{HINGE} and NeuInfer~\cite{NeuInfer} consider principal and subordinate structure information and represent each n-ary relational fact as a primary triple coupled with a set of its auxiliary descriptive r:v pair(s). Actually, no all n-ary relational facts have a primary triple. In a word, all these methods do not consider any type constraint.

\section{Problem Statement}
\label{sec:problem_statement}
We represent each n-ary relational fact as a set of its r:v pairs. As an example, the representation of the fact that $Marie\ Curie$ received $Nobel$ $Prize\ in\ Physics$ in $\textit{1903}$ together with $Henri\ Becquerel$ and $Pierre\ Curie$, is:
\begin{equation}
\begin{split}
	\{p&erson\!:\!Marie\ Curie, award\!:\!Nobel\ Prize\ in\ Physics, \\
	&point\ in\ time\!:\!\textit{1903}, together\ with\!:\!Henri\ Becquerel, \\
	&together\ with\!:\!Pierre\ Curie\}.
\nonumber
\end{split}
\end{equation}

Formally, given an n-ary relational fact with $n$ roles, and each role $r_i$ having $m_i$ role-values ($i=1, 2, \dots, n$), its representation is the following set of r:v pairs:
\begin{equation}
\begin{split}
	\{r&_1:v_1, \dots, r_1:v_{m_1}, r_2:v_{m_1+1}, \dots, r_2:v_{m_1+m_2}, \dots, \\
	&r_{n}:v_{m_1+m_2+\dots+m_{n-1}+1}, \dots, r_{n}:v_{m_1+m_2+\dots+m_{n}} \}.
\nonumber
\end{split}
\end{equation}
It is an n-ary relational fact of arity $(m_1+m_2+\dots+m_{n})$.

In what follows, without loss of generality, we exemplify one n-ary relational fact $Rel$ in the following form:
\begin{equation}
Rel=\{r_1:v_1, r_2:v_2, \dots, r_m:v_m\},
\nonumber
\end{equation}
where $m$ is the arity of $Rel$. The role set of $Rel$ is denoted as $R_{Rel}=\{r_1, r_2, \dots, r_m\}$, and $r_i$ may be the same to $r_j$ ($i,j=1,2,\dots,m, i\neq j$), as aforementioned; The role-value set of $Rel$ is denoted as $V_{Rel}=\{v_1, v_2, \dots, v_m\}$.

In this paper, we handle link prediction on n-ary relational data.
\begin{myDef}
	Link prediction on n-ary relational data is to predict the missing role/role-value in an n-ary relational fact, i.e., role prediction and role-value prediction. Formally, for the n-ary relational fact $Rel$, it predicts $r_i$/$v_i$ ($i=1,2,\dots,m$) given the remaining elements.
\end{myDef}
Take the above 5-ary relational fact for example, it may be to predict the role of the role-value $Pierre\ Curie$ or to predict the role-value of the role $award$, given the remaining information. In the following, it is transformed into estimating whether an n-ary relational fact is valid and then cast into a classification task. Actually, as each n-ary relational fact is represented as a set of its r:v pairs, the above classification task is operated on sets~\cite{Permutate_eq,Deep_sets,RNs}.

\begin{figure*}[!htb]
	\centering
	\includegraphics[width=6.8in]{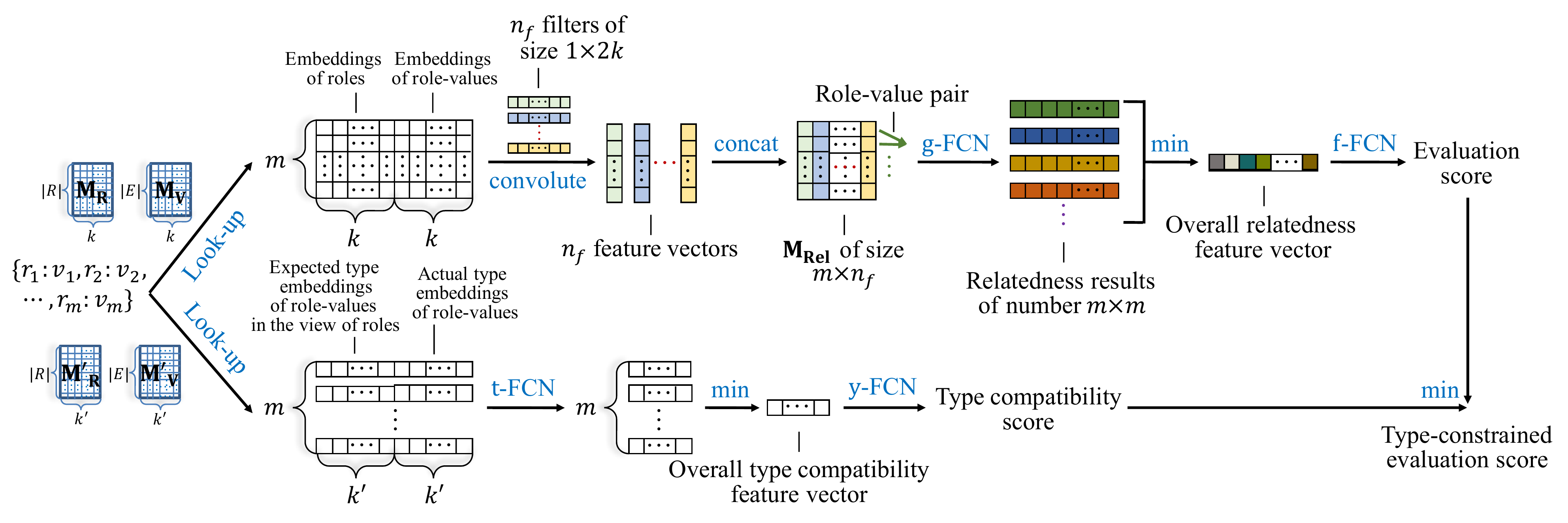}
	\vspace{-3mm}
	\vspace{-1mm}
	\caption{The framework of the proposed methods.}
	\label{fig:NaLP_framework}
	\vspace{-1mm}
	\vspace{-3mm}
\end{figure*}

\section{The NaLP Method}
\label{sec:NaLP_method}
\subsection{The Framework of NaLP}
\label{subsec:NaLP_framework}
We propose a link prediction method NaLP on n-ary relational data. It is designed based on two considerations. On the one hand, a role and the corresponding role-value are tightly linked to each other, thus should be bound together. On the other hand, as mentioned in Section~\ref{sec:problem_statement}, we aim to judge the validity of an n-ary relational fact, i.e., a set of r:v pairs. That is, for a set of r:v pairs, we need to determine whether it is able to form a valid n-ary relational fact. Intuitively, all the r:v pairs of a valid relational fact are closely related. And if all the r:v pairs from a set are closely related, then we assume that this set of r:v pairs has a high probability to constitute a valid n-ary relational fact. With these considerations in mind, the framework of NaLP is designed as illustrated in the upper part of Fig.~\ref{fig:NaLP_framework}, which consists of two key components, r:v pair embedding and relatedness evaluation, to generate the evaluation score of the input fact. For clarity, in the figure, we only exemplify the n-ary relational fact $Rel$, and in what follows, we illustrate its learning details.

For $Rel$, the embeddings of its roles in $R_{Rel}$ and role-values in $V_{Rel}$ are looked up from the embedding matrices $\mathbf{M_R}\in \mathbb{R}^{|R|\times k}$ of roles and $\mathbf{M_V}\in \mathbb{R}^{|V|\times k}$ of role-values, respectively, where $R$ is the set of all the roles in the dataset; $V$ is the set of all the role-values; $k$ is the dimension of embeddings. Following the convention, in what follows, embeddings are denoted with the same letters but in boldface. As depicted in Fig.~\ref{fig:NaLP_framework}, after passing these embeddings to the r:v pair embedding component, we get the embedding matrix of the $m$ r:v pairs, i.e., $Rel$ (see Section~\ref{subsec:rv_pair_embedding}). This resulting embedding matrix is fed into the relatedness evaluation component to compute the relatedness between the r:v pairs, then estimate the overall relatedness of all the r:v pairs, which is used to obtain the evaluation score (see Section~\ref{subsec:relatedness_evaluation}). Particularly, this framework equips NaLP with the ability to be permutation-invariant to the input order of the r:v pairs and deal with relational facts of different arities (see Section~\ref{subsec:look_into_NaLP}). Subsequently, the evaluation score is used to generate loss (see Section~\ref{subsec:model_training}).

\subsection{The R:V Pair Embedding Component}
\label{subsec:rv_pair_embedding}
This component aims to obtain the embeddings of the input r:v pairs. This process contains two sub-processes, i.e., feature learning of the r:v pairs (convolution is adopted, as it is a good method to learn features) and obtaining the embedding matrix of the r:v pairs, corresponding to ``convolute'' and ``concat'' in the upper part of Fig.~\ref{fig:NaLP_framework}, respectively.

\subsubsection{Feature Learning of the R:V Pairs}
As presented in Fig.~\ref{fig:NaLP_framework}, before convolution, the embeddings of the roles $r_i$ and the corresponding role-values $v_i$ ($i=1,$ $2, \dots,m$) in $Rel$ are concatenated, and form a matrix of dimension $m\times 2k$, where each r:v pair makes up a row.

Then, this concatenated matrix is passed to the convolution with the filter set $\Omega$ of $n_f$ filters. Since the dimensions of the concatenated embeddings of the r:v pairs are $1\times 2k$, we set the dimensions of the filters to the same value, i.e., $1\times 2k$ to capture their features. After convolution, we get $n_f$ results of dimension $m$. The Rectified Linear Units (ReLU) function~\cite{ReLU} is further applied to get $n_f$ feature vectors.

\subsubsection{The Embedding Matrix of the R:V Pairs}
We concatenate the $n_f$ feature vectors to form a matrix of dimension $m\times n_f$. Then, this matrix can be treated as the embeddings of the $m$ r:v pairs, and each row corresponds to the embedding of one r:v pair. Specifically, each entry of a row encodes the feature of this dimension.

Formally, the embedding matrix $\mathbf{M_{Rel}}\in \mathbb{R}^{m\times n_f}$ of the r:v pairs, i.e., $Rel$ is defined as:
\begin{equation}
\mathbf{M_{Rel}} = {\rm concat}\left(\Upsilon({\rm concat}(\mathbf{R_{Rel}}, \mathbf{V_{Rel}})\ast \Omega)\right),
\label{eq:M_Rel}
\end{equation}
where $\Upsilon$ is the ReLU function, i.e., $\Upsilon(x) = \max(0, x)$; $\ast$ denotes the convolution operator; ${\rm concat}(\mathbf{R_{Rel}}, \mathbf{V_{Rel}})$ is the aforementioned embedding concatenation process of the r:v pairs before convolution.

\subsection{The Relatedness Evaluation Component}
\label{subsec:relatedness_evaluation}
This component computes the relatedness between the r:v pairs, estimates the overall relatedness of all the r:v pairs, and obtains the evaluation score, corresponding to ``g-FCN'', ``min'' and ``f-FCN'' in the upper part of Fig.~\ref{fig:NaLP_framework}, respectively. Before introducing these sub-processes, let us see the principle behind this component.

\subsubsection{The Principle of Relatedness Evaluation}
\label{subsubsec:relatedness_evaluation_principle}
The principle goes deep into determining whether a set of r:v pairs is able to form a valid n-ary relational fact. As demonstrated in Section~\ref{subsec:NaLP_framework}, this is reduced to computing the overall relatedness of all the r:v pairs in the set. However, the overall relatedness of a set with more than two objects is intricate to measure. When the number of objects is large, corresponding to the fact of high arity, the evaluation is further complicated. Whereas, computing the relatedness between two objects is relatively much simpler. Thus, how to determine the tricky overall relatedness via the simple relatedness between the r:v pairs?

Straightforwardly, if all the r:v pairs form a closely related set, i.e., a valid n-ary relational fact, then every two pairs from the set are greatly related. Thus, for any two pairs, the values of their relatedness feature vector, measuring the relatedness in many different views, are all expected to be sufficiently large. That is, for each feature dimension, the minimum over this dimension of every two pairs is not allowed to be too small. Hence, based on this observation, we apply element-wise minimizing over the relatedness feature vectors of every two r:v pairs to approximately obtain the overall relatedness feature vector of all the r:v pairs. Accordingly, the relatedness evaluation component is designed to contain the estimation of the relatedness between the r:v pairs and the overall relatedness of all the r:v pairs, before computing the evaluation score.

\subsubsection{Relatedness Evaluation Between the R:V Pairs}
\label{subsubsec:pair-wise_relatedness_evaluation}
The relatedness between the r:v pairs is captured via a FCN with ReLU as its activation function. In particular, the widely used FCN is also adopted to infer how two objects are related and achieves marvelous performance in computer vision~\cite{Permutate_eq,RNs}. In this sub-process, it turns out that a one-layer FCN already works very well. Thus, only a fully connected layer is adopted.

Concretely, the embeddings of any two r:v pairs are concatenated and then fed into a fully connected layer with $n_{\text{gFCN}}$ nodes. The resulting vector is the relatedness feature vector, where each dimension indicates the relatedness of the input two r:v pairs in a certain respect.

\subsubsection{Overall Relatedness Evaluation}
\label{subsubsec:overall_relatedness_evaluation}
As discussed in Section~\ref{subsubsec:relatedness_evaluation_principle}, the overall relatedness feature vector of all the r:v pairs is simply approximated by element-wise minimizing over the relatedness feature vectors of every two r:v pairs. Thus, the overall relatedness feature vector $\mathbf{\mathcal{R}_{Rel}}$ of $Rel$ is defined as:
\begin{equation}
\vspace{-1mm}
\begin{split}
	\mathbf{\mathcal{R}_{Rel}} =\ &{\min}_{i,j=1}^m\big(\text{g-FCN}({\rm concat}([\mathbf{M_{Rel}}]_i, [\mathbf{M_{Rel}}]_j))\big)\\
	=\ &{\min}_{i,j=1}^m \Upsilon({\rm concat}([\mathbf{M_{Rel}}]_i, [\mathbf{M_{Rel}}]_j)\mathbf{W_{gFCN}}\\
	&+\mathbf{b_{gFCN}}),
\label{eq:relatedness_Rel}
\end{split}
\end{equation}
where ${\rm min}(\cdot)$ is the element-wise minimizing function; $[\mathbf{x}]_i$ is the $i$-th entry of $\mathbf{x}$; $\mathbf{W_{gFCN}}$ of dimension $2n_f\times n_{\text{gFCN}}$ and $\mathbf{b_{gFCN}}$ of dimension $n_{\text{gFCN}}$ are the weight matrix and bias vector of g-FCN, respectively. Each entry of the resulting $\mathbf{\mathcal{R}_{Rel}}$ implies the degree of the overall relatedness in terms of a certain feature corresponding to that dimension.

\subsubsection{The Evaluation Score}
Then, $\mathbf{\mathcal{R}_{Rel}}$ is used to generate the evaluation score. A one-layer FCN is also found to be sufficient and applied. In this way, the evaluation score $s_0(Rel)$ of $Rel$ is defined as:
\begin{equation}
\vspace{-1mm}
s_0(Rel) = \text{f-FCN}(\mathbf{\mathcal{R}_{Rel}}) = \mathbf{\mathcal{R}_{Rel}}\mathbf{W_{fFCN}}+b_{\text{fFCN}},
\label{eq:sr-Rel}
\end{equation}
where $\mathbf{W_{fFCN}}$ of dimension $n_{\text{gFCN}}\times 1$ and $b_{\text{fFCN}}$ are the weight matrix and bias variable of f-FCN, respectively.

\subsection{The Characteristics of NaLP}
\label{subsec:look_into_NaLP}
Note that, in m-TransH~\cite{m-TransH}, RAE~\cite{RAE}, HypE~\cite{HypE}, and GETD~\cite{GETD}, all relations along with their role sequences are predefined, and the inputs of the models are the instances of these relations, i.e., the sequences of ordered role-values. Since the $i$-th role-value of an input instance corresponds to the $i$-th role of its relation, the order of the input role-values is usually not allowed to be changed randomly. Whereas, each input of the proposed NaLP method is a set of r:v pairs, corresponding to an n-ary relational fact. Thus, the order of the objects in each input matters in m-TransH, RAE, HypE, and GETD, while it has no impact on NaLP. Actually, NaLP is permutation-invariant to the input order of the r:v pairs and is able to cope with facts of different arities.
\subsubsection{Permutation Invariance of the R:V Pairs}
Suppose that the r:v pairs of $Rel$ are input into NaLP in the original order with the indexes from $1$ to $m$, and a permutation $\rho$ to the indexes results in a new order with the indexes from $\rho(1)$ to $\rho(m)$. 

Thus, we simply explain the permutation invariance as:
\begin{itemize}
	\item In the process of r:v pair embedding, the embedding matrix $\mathbf{M_{\rho(Rel)}}$ of the permutated r:v pairs $\rho(Rel)$ is a permutation of the original $\mathbf{M_{Rel}}$, with the $\rho(i)$-th row of $\mathbf{M_{Rel}}$ placed in the $i$-th row of $\mathbf{M_{\rho(Rel)}}$, i.e., $[\mathbf{M_{\rho(Rel)}}]_i = [\mathbf{M_{Rel}}]_{\rho(i)}$.
	\item In the process of relatedness evaluation, the overall relatedness feature vector $\mathbf{\mathcal{R}_{\rho(Rel)}}$ of $\rho(Rel)$ is
\begin{equation}
\begin{split}
	\!\!\!\!\!\!\!\!\!\!\!\!\!\mathbf{\mathcal{R}}&_\mathbf{{\rho(Rel)}}\\
	\!\!\!\!\!\!\!\!\!\!\!\!\!= &\ {\min}_{i,j=1}^m\!\big(\text{g-FCN}({\rm concat}([\mathbf{M_{\rho(Rel)}}]_i, [\mathbf{M_{\rho(Rel)}}]_j))\big)\\
	\!\!\!\!\!\!\!\!\!\!\!\!\!= &\ {\min}_{i,j=1}^m\!\big(\text{g-FCN}({\rm concat}([\mathbf{M_{Rel}}]_{\rho(i)}, [\mathbf{M_{Rel}}]_{\rho(j)}))\big).
\end{split}
\end{equation}
That is, we obtain $m\times m$ relatedness feature vectors, the same to those in the original version, but with their order changed. Since ${\rm min}(\cdot)$ is permutation-equivariant, we have: $\mathbf{\mathcal{R}_{\rho(Rel)}} = \mathbf{\mathcal{R}_{Rel}}$.
	\item Then, the evaluation score $s_0(\rho(Rel))$ of $\rho(Rel)$ is
\begin{equation}
\!\!\!\!\!\!\!\!\!\!\!\!\!\!\!\!\!s_0(\rho(Rel)) \!=\! \text{f-FCN}(\mathbf{\mathcal{R}_{\rho(Rel)}}) \!=\! \text{f-FCN}(\mathbf{\mathcal{R}_{Rel}}) \!=\! s_0(Rel).\!\!\!\!\!\!\!\!\!
\end{equation}
	\item Therefore, NaLP is permutation-invariant to the input order of the r:v pairs.
\end{itemize}

Permutation invariance is also studied in DeepSets~\cite{Deep_sets}. It claimed that a function acting on sets must be permutation-invariant. However, it only sums up the features of each object in a set. \cite{Permutate_eq,RNs} further consider the pairwise interactions among objects. Similar to \cite{Permutate_eq,RNs}, we study these pairwise features, but on more complicated objects, each consisting of a role and its role-value.

\subsubsection{Feasibility to Different Arities}
Suppose that two batches of facts of arities $m$ and $m'$ are input into NaLP subsequently. In the process of r:v pair embedding, we get the embedding matrices of the r:v pairs with $m$ and $m'$ rows, respectively. In the process of relatedness evaluation, correspondingly, we obtain $m\times m$ and $m'\times m'$ relatedness feature vectors. As ${\rm min}(\cdot)$ in Equation~\eqref{eq:relatedness_Rel} returns one vector among all these results of number $m\times m$ or $m'\times m'$, the number of results makes no difference. Thus, NaLP is able to deal with facts of different arities.

\subsection{Model Training}
\label{subsec:model_training}
\subsubsection{The Loss Function}
As above described, we obtain the evaluation score $s_0(Rel)$ of $Rel$. Following ComplEx~\cite{ComplEx}, the loss of $Rel$ is:
\begin{equation}
\mathcal{L}(Rel) = \log\! \left(1+e^{-I_{Rel}s_0(Rel)} \right)\!,
\label{eq:loss-Rel}
\end{equation}
where $I_{Rel}=1$, if $Rel$ is valid, and $I_{Rel}=-1$, otherwise.

It is straightforward to optimize NaLP with the standard backpropagation. The stochastic optimization method Adam~\cite{Adam} with learning rate $\lambda$ is used as the optimizer.

\subsubsection{The Training Process}
\label{subsubsec:training_process}
Algorithm~\ref{alg:train_algorithm} presents the training process of NaLP.
\begin{algorithm}
	\caption{The training process of NaLP.}
	\begin{algorithmic}[1]
		\REQUIRE Training set $T$, role set $R$, role-value set $V$, max number of epochs $nepoch$, embedding dimension $k$, batch size $\beta$, number of filters $n_f$ in $\Omega$, the hyper-parameter $n_{\text{gFCN}}$ of $\mathbf{W_{gFCN}}$.
		\ENSURE $\mathbf{M_R}$ and $\mathbf{M_V}$, as well as the parameters of NaLP.
		\STATE $T\leftarrow$ group $T$ by the arities of the facts;~\label{group}
		\STATE $T\leftarrow$ sort the groups in $T$ by arities in ascending order;~\label{sort}
		\STATE \textbf{Initialize} $\mathbf{M_R}, \mathbf{M_V}\leftarrow{\rm uniform}(-\frac{1}{\sqrt{k}}, \frac{1}{\sqrt{k}})$;\\
		\ \ \ \ \ \ \ \ \ \ \ \ \ \ \ \ \ $\Omega\leftarrow{\rm truncated\_normal}(0.0, 0.1)$;\\
		\ \ \ \ \ \ \ \ \ \ \ \ \ \ \ \ \ $\mathbf{W_{gFCN}}, \mathbf{W_{fFCN}}\leftarrow{\rm xavier\_initializer()}$;\\
		\ \ \ \ \ \ \ \ \ \ \ \ \ \ \ \ \ $\mathbf{b_{gFCN}}, b_{\text{fFCN}}\leftarrow{\rm zeros\_initializer()}$;~\label{init_ER}
		\REPEAT
		\FOR{each group $T'\in T$~\label{train_begin}}
		\FOR{$j=1$ to $\left\lceil\frac{|T'|}{\beta}\right\rceil$}
		\STATE $S\textsuperscript{+}\leftarrow$ sample$(T', \beta)$;~\label{sample}\ \COMMENT{sample $\beta$ training facts}
		\STATE $S\textsuperscript{-}\leftarrow$ neg\_sample$(S\textsuperscript{+}, \beta)$;~\label{neg_sample}\ \COMMENT{get $\beta$ negative ones}
		\STATE $loss \leftarrow 0$;
		\FOR{$\forall Rel\in S\textsuperscript{+}\cup S\textsuperscript{-}$}
		\STATE $\mathbf{R_{Rel}}\leftarrow\ $look-up$(\mathbf{M_R}, R_{Rel})$;~\label{look_up_roles}\ \COMMENT{look up $R_{Rel}$}
		\STATE $\mathbf{V_{Rel}}\leftarrow\ $look-up$(\mathbf{M_V},V_{Rel})$;~\label{look_up_values}\ \COMMENT{look up $V_{Rel}$}
		\STATE $\mathbf{M_{Rel}}\leftarrow\ $get the embedding matrix of the r:v pairs following Equation~\eqref{eq:M_Rel};~\label{get-M_Rel}
		\STATE $\mathbf{\mathcal{R}_{Rel}}\leftarrow\ $compute the overll relatedness feature vector of $Rel$ following Equation~\eqref{eq:relatedness_Rel};~\label{get-relatedness_Rel}
		\STATE $s_0(Rel)\leftarrow\ $generate the evaluation score following Equation~\eqref{eq:sr-Rel};~\label{get-s-Rel}
		\STATE $\mathcal{L}(Rel)\leftarrow\ $compute the loss of $Rel$ following Equation~\eqref{eq:loss-Rel};~\label{compute_loss}
		\STATE $loss\leftarrow loss + \mathcal{L}(Rel)$;~\label{add_loss}
		\ENDFOR
		\STATE Update the embeddings of the roles and role-values in $S\textsuperscript{+}\cup S\textsuperscript{-}$ (i.e., the related rows of $\mathbf{M_R}$ and $\mathbf{M_V}$), the filters in $\Omega$, $\mathbf{W_{gFCN}}$, $\mathbf{W_{fFCN}}$, $\mathbf{b_{gFCN}}$, as well as $b_{\text{fFCN}}$ via $\bigtriangledown {loss}$;~\label{param_update}
		\ENDFOR
		\ENDFOR~\label{train_end}
		\UNTIL{the result on the validation set converges}
	\end{algorithmic}
\label{alg:train_algorithm}
\end{algorithm}

Before training, the training set is reorganized into several groups, each of which keeps the facts of the same arities (Line~\ref{group}). In order to guarantee that NaLP masters basic facts of low arities before learning facts of high arities, we sort the above resulting groups by their arities in ascending order (Line~\ref{sort}). Similar operations are applied to train the model on the path queries of length one and then all the path queries~\cite{Guu_traverseKG}. Subsequently, the embedding matrices $\mathbf{M_R}$ and $\mathbf{M_V}$ are randomly initialized from uniform distribution, respectively, with $-\frac{1}{\sqrt{k}}$ as minimum and $\frac{1}{\sqrt{k}}$ as maximum; Truncated normal distribution with 0.0 as mean and 0.1 as standard deviation is adopted as initializer to initialize all the filters in $\Omega$ \cite{ConvKB}; $\mathbf{W_{gFCN}}$ and $\mathbf{W_{fFCN}}$ are randomly initialized via xavier initializer~\cite{xavier_initializer}, respectively; $\mathbf{b_{gFCN}}$ and $b_{\text{fFCN}}$ are randomly initialized with zeros (Line~\ref{init_ER}).

During training, Lines~\ref{train_begin}-\ref{train_end} are repeated until the result on the validation set converges. In each epoch, $\left\lceil |T'|/\beta\right\rceil$ batches of training facts are sampled from each training group $T'$, where $\lceil \cdot\rceil$ is the ceiling function. For each selected training fact, similar to the negative sampling method adopted in TransE~\cite{TransE}, we randomly replace one of its role-values with a random role-value that the corresponding role holds with probability $|V|/(|V|+|R|)$, or one of its roles is replaced by a random role with probability $|R|/(|V|+|R|)$, to obtain a negative sample not contained in the dataset (Line~\ref{neg_sample}). After that, for each sampled n-ary relational fact or negative one $Rel$, the embeddings of its roles and role-values are looked up from $\mathbf{M_R}$ and $\mathbf{M_V}$, respectively (Lines~\ref{look_up_roles} and \ref{look_up_values}). Then, they are fed into the r:v pair embedding component to obtain the embedding matrix of the r:v pairs (Line~\ref{get-M_Rel}). After passing this embedding matrix to the relatedness evaluation component, we obtain the overall relatedness feature vector $\mathbf{\mathcal{R}_{Rel}}$ and the evaluation score $s_0(Rel)$ of $Rel$ (Lines~\ref{get-relatedness_Rel} and \ref{get-s-Rel}). Subsequently, based on $s_0(Rel)$, loss is computed and added to the loss of the current batch (Lines~\ref{compute_loss} and \ref{add_loss}). Finally, parameters are updated in batch mode (Line~\ref{param_update}).

Specifically, batch normalization~\cite{BN} is applied after convolution in the r:v pair embedding component to stabilize and accelerate the training process.

\section{tNaLP, NaLP\textsuperscript{+}, and tNaLP\textsuperscript{+}}
\label{sec:NaLP_extend_methods}

\subsection{tNaLP}
\label{subsec:tNaLP}
NaLP does not consider any type constraint to estimate the validity of an n-ary relational fact. Actually, for a valid n-ary relational fact, the expected type of each role-value in the view of the corresponding role should be compatible with the actual type of the role-value. Take the fact in Section~\ref{sec:problem_statement} for example, the roles $person$, $award$, $point\ in\ time$, and $together\ with$ expect the corresponding role-values of type $human$, $award$, $time$, and $human$, respectively. The actual types of the corresponding role-values in the mentioned fact do meet this requirement. Thus, with these considerations in mind, we further introduce these type constraints and extend NaLP to tNaLP. Not relying on any explicit external type supervision, tNaLP learns type embeddings in the training process. The framework of tNaLP is illustrated in Fig.~\ref{fig:NaLP_framework}, where the upper part obtains the evaluation score of the input fact, corresponding to NaLP, and the lower part obtains its type compatibility score. They are finally combined to obtain the type-constrained evaluation score.

Similar to the relatedness evaluation component in Section~\ref{subsec:relatedness_evaluation}, before obtaining the type compatibility score, we compute the type compatibility of each r:v pair and estimate the overall type compatibility of all the r:v pairs.

\subsubsection{Type Compatibility Evaluation of each R:V Pair}
Following Section~\ref{subsubsec:pair-wise_relatedness_evaluation}, we adopt a one-layer FCN with ReLU as its activation function to estimate the type compatibility, denoted as ``t-FCN'' in Fig.~\ref{fig:NaLP_framework}.

Specifically, for each r:v pair of $Rel$, the expected type embedding of the role-value in the view of its role and the actual type embedding of the role-value are looked up from the expected type embedding matrix $\mathbf{M'_R}\in \mathbb{R}^{|R|\times k'}$ and the type embedding matrix $\mathbf{M'_V}\in \mathbb{R}^{|V|\times k'}$, respectively, where $k'$ is the dimension of type embeddings. They form the expected type embedding matrice $\mathbf{R'_{Rel}}$ and the actual type embedding matrice $\mathbf{V'_{Rel}}$, respectively. Then, each row of $\mathbf{R'_{Rel}}$ and the corresponding row of $\mathbf{V'_{Rel}}$ are concatenated to form a vector of dimension $2k'$, and fed into a fully connected layer with $n_{\text{tFCN}}$ nodes. Each resulting type compatibility feature vector indicates the type compatibility of the corresponding role and role-value.

\subsubsection{Overall Type Compatibility Evaluation}
As Section~\ref{subsubsec:overall_relatedness_evaluation} does, the overall type compatibility feature vector $\mathbf{C_{Rel}}$ of $Rel$ is approximated by element-wise minimizing over the type compatibility feature vectors of all the r:v pairs. Therefore, $\mathbf{C_{Rel}}$ is defined as:
\begin{equation}
\vspace{-1mm}
\begin{split}
	\mathbf{C_{Rel}} \!=\ &{\min}_{i,j=1}^m\big(\text{t-FCN}({\rm concat}([\mathbf{R'_{Rel}}]_i, [\mathbf{V'_{Rel}}]_j))\big)\\
	=\ &{\min}_{i,j=1}^m \Upsilon({\rm concat}([\mathbf{R'_{Rel}}]_i, [\mathbf{V'_{Rel}}]_j)\mathbf{W_{tFCN}}\\
	&+\mathbf{b_{tFCN}}),
\label{eq:tcompatibility_Rel}
\end{split}
\end{equation}
where $\mathbf{W_{tFCN}}$ of dimension $2k'\times n_{\text{tFCN}}$ and $\mathbf{b_{tFCN}}$ of dimension $n_{\text{tFCN}}$ are the weight matrix and bias vector of t-FCN, respectively.

\subsubsection{The Type Compatibility Score and Type-constrained Evaluation Score}
Subsequently, $\mathbf{C_{Rel}}$ is fed into a one-layer FCN to obtain the type compatibility score $s_t(Rel)$ of $Rel$, denoted as ``y-FCN'' in Fig.~\ref{fig:NaLP_framework}, i.e.,
\begin{equation}
\vspace{-1mm}
s_t(Rel) = \text{y-FCN}(\mathbf{C_{Rel}}) = \mathbf{C_{Rel}}\mathbf{W_{yFCN}}+b_{\text{yFCN}},
\label{eq:st-Rel}
\end{equation}
where $\mathbf{W_{yFCN}}$ of dimension $n_{\text{tFCN}}\times 1$ and $b_{\text{yFCN}}$ are the weight matrix and bias variable of y-FCN, respectively.

If $Rel$ is valid, then its evaluation score $s_0(Rel)$ from NaLP and its type compatibility score $s_t(Rel)$ are expected to be large. That is, the minimum of them is encouraged to be large. Thus, we adopt the minimizing operation to obtain the type-constrained evaluation score $s(Rel)$ of tNaLP:
\begin{equation}
\vspace{-1mm}
	s(Rel) = \min(s_0(Rel), s_t(Rel)).
\label{eq:s-Rel}
\end{equation}

\subsection{NaLP\textsuperscript{+} and tNaLP\textsuperscript{+}}
\label{subsec:NaLP_plus}
In NaLP and tNaLP, the conventional negative sampling method (see Section~\ref{subsubsec:training_process}) similar to that in TransE~\cite{TransE} is adopted. Although this negative sampling mechanism that replaces only one element of a fact is effective on binary relational datasets~\cite{TransE}, it is not the case on n-ary relational datasets. Actually, it only pushes the model to distinguish a valid element from other elements in the dataset. However, it does not encourage the model to distinguish the r:v pairs in one n-ary relational fact from those in others. 

Therefore, in this paper, we propose a more reasonable negative sampling mechanism to take the distinguishment among n-ary relational facts into consideration. Upon each n-ary relational fact of arity $m$, the negative sampling process under this new mechanism is described as follows:
\begin{enumerate}[(1)]
	\item Decide to replace one role-value/role or r:v pair(s) randomly. If replace one role-value/role, go to Step~(\ref{item:2}), else go to Step~(\ref{item:3});
	\item Randomly replace one of the role-values or roles in the fact following Section~\ref{subsubsec:training_process};~\label{item:2}
	\item Decide the random number $n_{neg}\in (0, m)$ of r:v pairs to be replaced, then choose $n_{neg}$ r:v pairs from the training set randomly, which may belong to different facts, to replace the $n_{neg}$ random r:v pairs of the given fact.~\label{item:3}
\end{enumerate}

Under this more reasonable negative sampling mechanism, NaLP and tNaLP are extended to NaLP\textsuperscript{+} and tNaLP\textsuperscript{+}, respectively. Note that, since the element-wise minimizing function in the overall type compatibility evaluation is permutation-equivariant and the proposed more reasonable negative sampling mechanism only affects the negative sampling process, tNaLP, NaLP\textsuperscript{+}, and tNaLP\textsuperscript{+} are permutation-invariant to the input order of the r:v pairs and are able to handle facts of different arities.

\section{Experiments}
\label{sec:experiments}
\subsection{Datasets}
\label{subsec:datasets}
\begin{table*}[htbp]
	\setlength{\tabcolsep}{0.7em}
	\centering
	\caption{The statistics of the two experimental datasets, JF17K and WikiPeople.}
	\vspace{-1mm}
	\vspace{-1mm}
	\begin{tabular}{c|c|ccc|ccc|ccc|ccc}
		\toprule[1pt]
		\multirow{2}{*}{Dataset} &\multirow{2}{*}{$|V|$} &\multicolumn{3}{c|}{$\#Relation$} &\multicolumn{3}{c|}{$\#Train$} &\multicolumn{3}{c|}{$\#Valid$} &\multicolumn{3}{c}{$\#Test$}\\
		\cline{3-14} & &Binary &N-ary &Overall &Binary &N-ary &Overall &Binary &N-ary &Overall &Binary &N-ary &Overall\\
		\midrule[0.5pt]
		JF17K &29,177 &191 &136 &327 &36,293 &25,618 &61,911 &9,271 &6,551 &15,822 &10,758 &14,157 &24,915\\
		WikiPeople &47,765 &150 &557 &707 &270,179 &35,546 &305,725 &33,845 &4,378 &38,223 &33,890 &4,391 &38,281\\
		\bottomrule[1pt]
	\end{tabular}
	\vspace{-1mm}
	\vspace{-1mm}
	\label{table:datasets}	
\end{table*}
\begin{table*}[htb]
	\setlength{\tabcolsep}{1em}
	\centering
	\caption{Comparison results of NaLP, tNaLP, NaLP\textsuperscript{+}, and tNaLP\textsuperscript{+} on role-value prediction in terms of MRR and Hits@\{1, 3, 10\} on JF17K and WikiPeople.}
	\vspace{-1mm}
	\vspace{-1mm}
	\makebox[\columnwidth][c]{
		\begin{tabular}{c|cccc|cccc}
			\toprule[1pt]
			\multirow{2}{*}{Method} &\multicolumn{4}{c|}{JF17K} &\multicolumn{4}{c}{WikiPeople}\\
			\cline{2-9} &MRR &Hits@1 &Hits@3 &Hits@10 &MRR &Hits@1 &Hits@3 &Hits@10\\
			\midrule[0.5pt]
			NaLP &0.386 &0.317 &0.413 &0.517     &0.338 &0.272 &0.364 &0.466\\
			tNaLP &0.384 &0.315 &0.411 &0.513    &\textbf{0.350} &\textbf{0.288} &\textbf{0.377} &0.471\\
			NaLP\textsuperscript{+} &0.432 &0.352 &0.465 &0.582    &0.338 &0.268 &0.365 &\textbf{0.475}\\
			tNaLP\textsuperscript{+} &\textbf{0.449} &\textbf{0.370} &\textbf{0.484} &\textbf{0.598}    &0.339 &0.269 &0.369	 &0.473\\
			\bottomrule[1pt]
		\end{tabular}
	}
	\vspace{-4mm}
	\label{table:extend_methods_comparison}
\end{table*}

We conduct experiments on two datasets. The first one is the public n-ary relational dataset JF17K\footnote{As the original version has no validation set, we use a new one, where 20\% facts in the training set is randomly selected to form the validation set.}~\cite{m-TransH}, derived from the popular KG Freebase~\cite{Freebase}. For all its facts, the role sequences are equal to the predefined standard ones and the corresponding role-values are existing and known. This is usually not the case. Thus, we build a relatively more practical dataset WikiPeople, which is the second experimental dataset. It is constructed as follows:  
\begin{itemize}
	\item We download the Wikidata dump\footnote{https://archive.org/details/wikibase-wikidatawiki-20171120\label{footnote:wikidata_dump}} and extract the facts concerning entities of type $human$.
	\item Then, these facts are further denoised. For example, facts containing element related to image are filtered out, and facts containing element in $\{unknown\ value, no\ values\}$ are removed.
	\item Subsequently, we select the subsets of elements which have at least $30$ mentions. The facts related to these elements are kept. Each fact is further parsed into a set of its r:v pairs.
	\item The remaining facts are randomly split into training set, validation set, and test set by a percentage of 80\%, 10\%, and 10\%, respectively. 
\end{itemize}

Actually, WikiPeople is relatively more flexible, where data incompleteness, insertion, and update are universal. Taking ``$Marie\ Curie$ received $Nobel\ Prize\ in\ Chemistry$ in \textit{1911}'' as an example, its role sequence is $[person, award,$ $point\ in\ time]$. Then, in the view of this fact, the following similar facts in Wikidata\footnote{https://www.wikidata.org/wiki/Q7186} correspond to the three cases of data incompleteness, insertion, and update, respectively:
\begin{itemize}
	\item $Marie\ Curie$ received $Willard\ Gibbs\ Award$. 
	\item $Marie\ Curie$ received $Matteucci\ Medal$ in $\textit{1904}$ with \underline{$Pierre$} \underline{$Curie$}.
	
	$Marie\ Curie$ received $Nobel$ $Prize\ in\ Physics$ in $\textit{1903}$ together with \underline{$Henri\ Becquerel$} and \underline{$Pierre$} \underline{$Curie$}.  
	\item $Marie\ Curie$ received $Davy\ Medal$ with \underline{$Pierre$} \underline{$Curie$}.
\end{itemize}
In the first example, the role $point\ in\ time$ and its role-value are missing. Thus, it is incomplete. In the second examples, one or two new roles $together\ with$ should be inserted into $[person, award, point\ in\ time]$. In the third one, the role $point\ in\ time$ is missing but a new role $together\ with$ is inserted. Thus, we call such a case as an update one. Fortunately, our representation form of n-ary relational facts, representing each one as a set of its r:v pairs, is able to cope with all these cases elegantly.

In the real scenario, since there is possible missing in knowledge extraction, and knowledge also grows and updates rapidly with many new contents flowing in, in the dataset derived from one snapshot, the above phenomena of data incompleteness, insertion, and update are inevitable. Thus, the developed WikiPeople is practical. Notably, the role-values from the continuous domain (e.g., role-values corresponding to point in time, date of birth, date of death, etc.) are tackled like others from the discrete domain.

The detailed statistics of the two experimental datasets are displayed in Table~\ref{table:datasets}, where $\#Relation$, $\#Train$, $\#Valid$, and $\#Test$ are the sizes of the relation set, the training set, the validation set, and the test set, respectively. Here, $\#Relation$ is counted according to RAE~\cite{RAE} to have a better understanding of the two datasets.

\subsection{Metrics and Experimental Settings}
\label{subsec:metrics_experiment_settings}
As for metrics, we adopt the standard Mean Reciprocal Rank (MRR) and Hits@$N$. These metrics are computed in a similar way of binary relational dataset~\cite{TransE}. For each test fact, one of its roles/role-values is removed and replaced by all the other roles/role-values in $R$/$V$. These corrupted facts and the test fact are fed into the proposed methods to obtain the (type-constrained) evaluation scores. Then, they are sorted according to their (type-constrained) evaluation scores in descending order. The rank of the test fact is finally stored. This whole procedure is repeated for all the other roles/role-values of the test fact. Thus, MRR is the average of these reciprocal ranks and Hits@$N$ is the percentage of ranks which are less than or equal to $N$. The traditional mean rank (the average of these ranks) is not adopted as a metric, since it is sensitive to outliers~\cite{HolE}. For these chosen metrics, the higher the value of MRR/Hits@$N$, the better the performance. In the test process, corrupted facts that may also be valid, i.e., exist in the training/validation/test set, are discarded before sorting the facts.

For the proposed methods, the reported results are given under the best set of hyper-parameters on the validation set after grid search on the following ranges, by reference to ConvKB~\cite{ConvKB}: The embedding dimension $k\in \{50, 100\}$, the batch size $\beta\in \{128, 256\}$, the learning rate $\lambda \in \{5e^{-6}, 1e^{-5}, 5e^{-5}, 1e^{-4}, 5e^{-4}, 1e^{-3}\}$, the number of filters $n_f$ of the convolution in $\{50, 100, 200, 400, 500\}$, the hyper-parameter $n_{\text{gFCN}}$ of $\mathbf{W_{gFCN}}$ in $\{50, 100, 200, 400, 500, 800,$ $1000, 1200\}$, the type embedding dimension $k'\in \{5, 10,$ $20, 30, 40\}$, and the hyper-parameter $n_{\text{tFCN}}$ of $\mathbf{W_{tFCN}}$ in $\{5, 10, 20, 40, 50, 80, 100, 150, 200\}$.

\subsection{The Effectiveness of tNaLP, NaLP\textsuperscript{+}, and tNaLP\textsuperscript{+}}
\label{subsec:NaLP_extend_effectiveness}
In this section, we analyze the effectiveness of the extended methods to further select methods to compare with the baselines in the next section. As role prediction is much simpler due to the much smaller role set, the above decision is made on more difficult role-value prediction. Table~\ref{table:extend_methods_comparison} demonstrates the experimental results.

 Specifically, on JF17K, tNaLP\textsuperscript{+} improves NaLP by 0.063, 5.3\%, 7.1\%, and 8.1\% on MRR and Hits@\{1, 3, 10\}, respectively. It verifies the superiority of combining type constraints unsupervisedly and the newly proposed negative sampling mechanism. This negative sampling mechanism, which replaces a random number of r:v pairs, encourages NaLP\textsuperscript{+} and tNaLP\textsuperscript{+} to distinguish not only a valid element from other elements in the dataset but also the r:v pairs in one n-ary relational fact from those in others, and results in much better performance.
 
 \begin{table}[htb]
	\setlength{\tabcolsep}{1em}
	\centering
	\caption{Comparison results of NaLP, tNaLP, NaLP\textsuperscript{+}, and tNaLP\textsuperscript{+} on role-value prediction in terms of MRR and Hits@\{1, 3, 10\} on WikiPeople-n.}
	\vspace{-1mm}
	\vspace{-1mm}
	\makebox[\columnwidth][c]{
		\begin{tabular}{c|cccc}
			\toprule[1pt]
			Method &MRR &Hits@1 &Hits@3 &Hits@10\\
			\midrule[0.5pt]
			NaLP &0.197 &0.127 &0.219 &0.338\\
			tNaLP &0.201 &0.132 &0.221 &0.337\\
			NaLP\textsuperscript{+} &0.217 &0.141 &0.240 &0.365\\
			tNaLP\textsuperscript{+} &\textbf{0.225} &\textbf{0.148} &\textbf{0.250} &\textbf{0.376}\\
			\bottomrule[1pt]
		\end{tabular}
	}
	\vspace{-4mm}
	\label{table:extend_methods_WikiPeople-n}
\end{table}

\begin{table*}[htb]
	\setlength{\tabcolsep}{0.9em}
	\centering
	\caption{Comparison results of NaLP, tNaLP, NaLP\textsuperscript{+}, and tNaLP\textsuperscript{+} on role-value prediction in terms of MRR and Hits@\{1, 3, 10\} on WikiPeople-0bi, WikiPeople-50bi, and WikiPeople-100bi.}
	\vspace{-1mm}
	\vspace{-1mm}
	\makebox[\columnwidth][c]{
		\begin{tabular}{c|cccc|cccc|cccc}
			\toprule[1pt]
			\multirow{2}{*}{Method} &\multicolumn{4}{c|}{WikiPeople-0bi} &\multicolumn{4}{c|}{WikiPeople-50bi}&\multicolumn{4}{c}{WikiPeople-100bi}\\
			\cline{2-13} &MRR &Hits@1 &Hits@3 &Hits@10 &MRR &Hits@1 &Hits@3 &Hits@10 &MRR &Hits@1 &Hits@3 &Hits@10\\
			\midrule[0.5pt]
			NaLP &0.300 &0.208 &0.339 &0.480 &0.220 &\textbf{0.152} &0.240 &0.354 &0.405 &\textbf{0.354} &0.425 &0.501\\
			tNaLP &0.314 &0.219 &0.356 &0.495 &0.191 &0.126 &0.212 &0.318 &\textbf{0.409} &0.353 &\textbf{0.433} &0.510\\
			NaLP\textsuperscript{+} &0.315 &0.218 &0.358 &0.496 &\textbf{0.229} &\textbf{0.152} &\textbf{0.254} &\textbf{0.376} &0.396 &0.341 &0.416 &0.498\\
			tNaLP\textsuperscript{+} &\textbf{0.320} &\textbf{0.223} &\textbf{0.363} &\textbf{0.500} &0.218 &0.144 &0.243 &0.363 &0.407 &0.349 &0.432 &\textbf{0.516}\\
			\bottomrule[1pt]
		\end{tabular}
	}
	\vspace{-4mm}
	\label{table:extend_methods_WikiPeople-xbi}
\end{table*}

On WikiPeople, tNaLP performs better than NaLP, which indicates the necessity of type constraints. Although we introduce type constraints in an unsupervised manner, it does work. However, from the results of NaLP\textsuperscript{+} and tNaLP\textsuperscript{+}, we can observe that the new negative sampling mechanism does not present its effectiveness. It is probably because of the much higher percentage of binary relational facts on WikiPeople. To further verify this conjecture, we derive a new dataset, WikiPeople-n, from WikiPeople. Detailedly, based on WikiPeople, we keep all the n-ary relational facts, and randomly remove some binary relational facts to obtain the same percentage of binary and n-ary categories as in the training set on JF17K. The experimental results on WikiPeople-n are presented in Table~\ref{table:extend_methods_WikiPeople-n}. From the results, we can observe that NaLP\textsuperscript{+} and tNaLP\textsuperscript{+} are much better than NaLP and tNaLP, respectively, which confirms the above conjecture. However, the overall performance on WikiPeople-n is worse than that on JF17K and WikiPeople. WikiPeople, where the phenomena of data incompleteness, insertion, and update are universal, is more practical and difficult than JF17K (see Section~\ref{subsec:datasets}). With the removal of many binary relational facts from WikiPeople, the above phenomena get severer, and thus WikiPeople-n becomes more difficult to deal with than JF17K and WikiPeople.

What about further vary the percentage of binary relational facts? To go deep into it, we vary the percentage (0\%, 50\%, and 100\%) of binary relational facts in WikiPeople to get three datasets, WikiPeople-0bi, WikiPeople-50bi, and WikiPeople-100bi, respectively. The experimental results are demonstrated in Table~\ref{table:extend_methods_WikiPeople-xbi}. It can be observed that tNaLP\textsuperscript{+} gets best performance on WikiPeople-0bi, NaLP\textsuperscript{+} on WikiPeople-50bi, and tNaLP on WikiPeople-100bi. From these results and the results in Tables~\ref{table:extend_methods_comparison}-\ref{table:extend_methods_WikiPeople-n}, we find that on the datasets of dominant binary relational facts (e.g., Wikipeople and WikiPeople-100bi), the new negative sampling mechanism is not so effective; Otherwise, it is superior.

\begin{figure}[!htb]
\vspace{-3mm}
	\centering
	\includegraphics[width=3.30in]{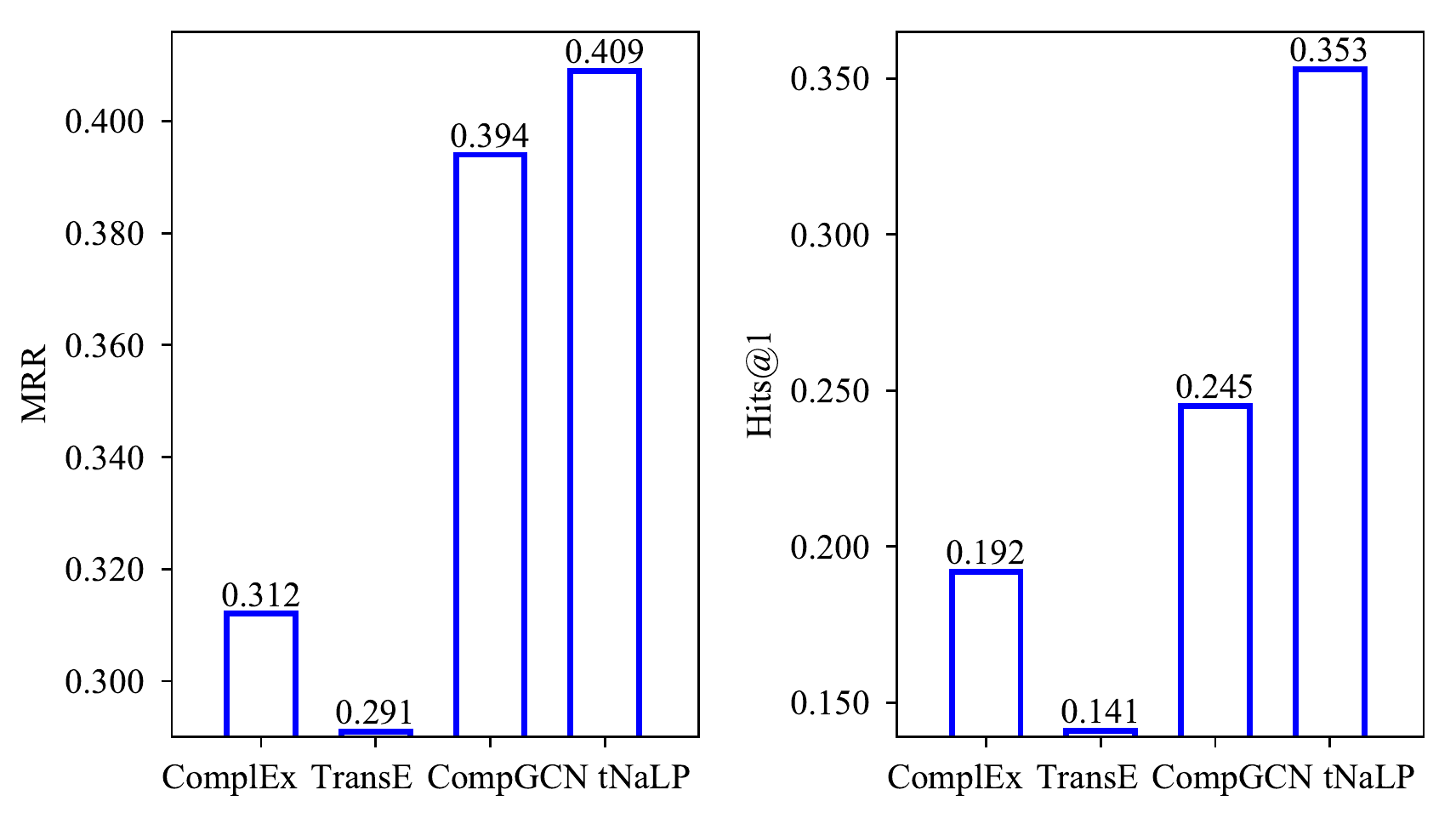}
	\vspace{-4mm}
	\caption{Comparison results of tNaLP and the representative KG methods in terms of MRR and Hits@1 on WikiPeople-100bi.}
	\label{fig:WikiPeople-100bi_comparison}
	\vspace{-3mm}
\end{figure}

Since WikiPeople-100bi is a KG dataset, we compare the proposed tNaLP method with the representative tensor/matrix based KG method ComplEx~\cite{ComplEx}, translation based one TransE~\cite{TransE}, and neural network based one CompGCN~\cite{CompGCN} in terms of fine-grained MRR and Hits@1 in Fig.~\ref{fig:WikiPeople-100bi_comparison}. We observe that tNaLP outperforms the representative KG methods significantly and thus is also superior on KG dataset.
 
\begin{figure}[!htb]
	\vspace{-1mm}
	\centering
	\includegraphics[width=2.75in]{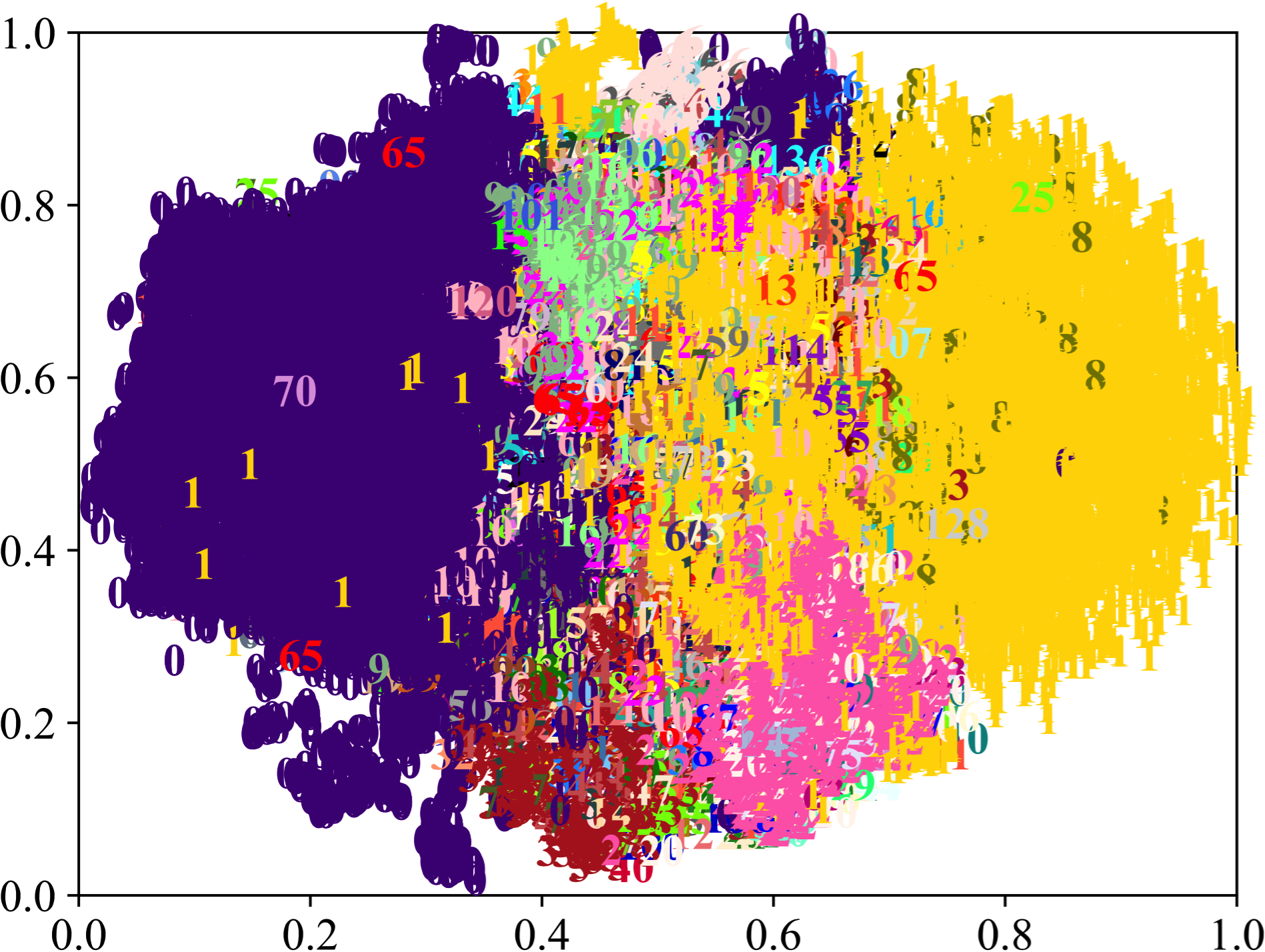}
	\vspace{-3mm}
	\caption{The visualization of the type embeddings of the role-values on WikiPeople.}
	\label{fig:type_embed_WikiPeople}
	\vspace{-2mm}
\end{figure}

\begin{figure}[!htb]
	\centering
	\includegraphics[width=2.85in]{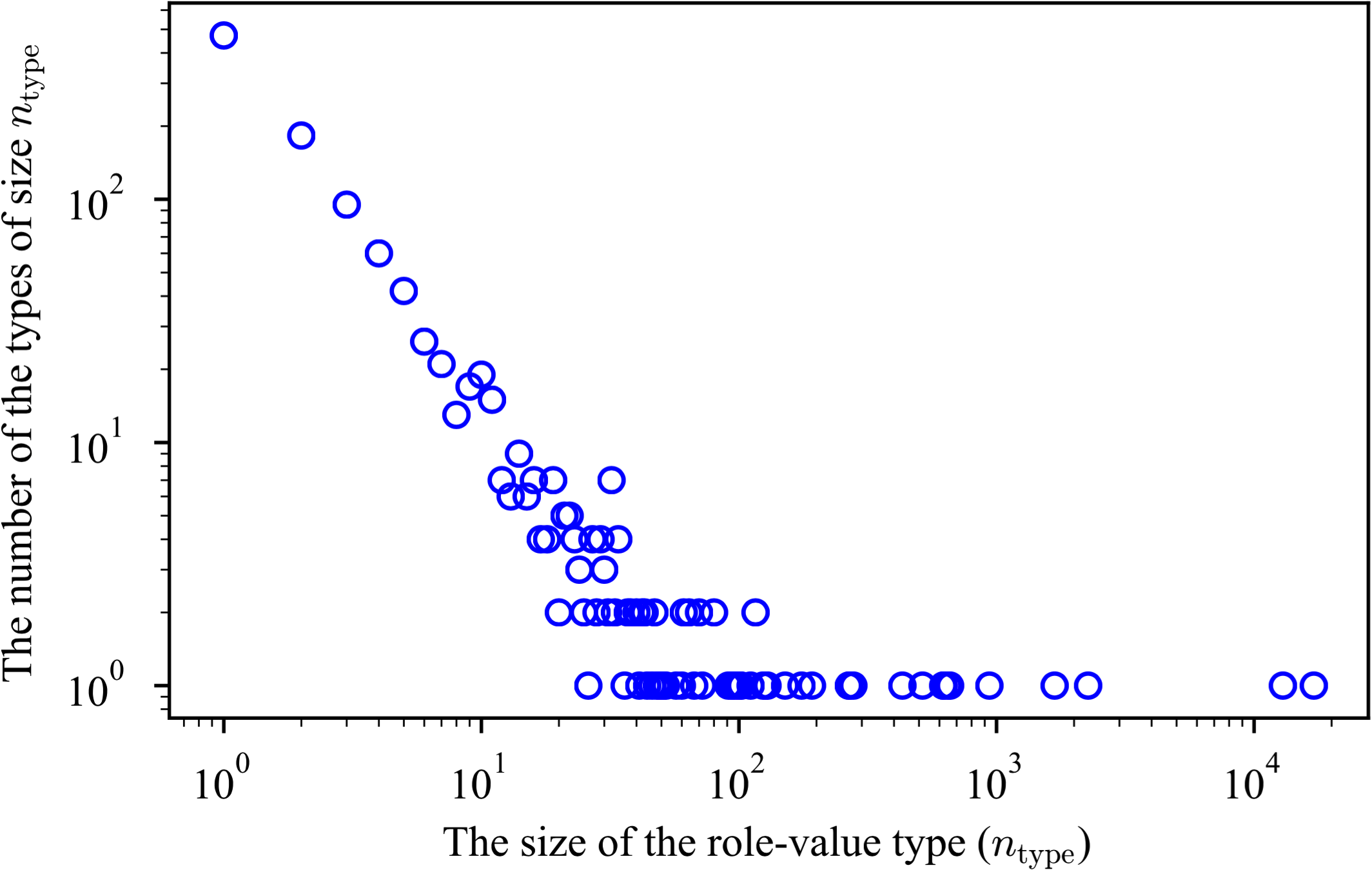}  
	\vspace{-3mm}
	\caption{The statistics of the role-value types on WikiPeople (in logarithm).}
	\label{fig:type_stat_WikiPeople}
	\vspace{-1mm}
	\vspace{-3mm}
\end{figure}

Notably, on all the datasets, tNaLP does not gain significant performance improvement, as compared to NaLP, especially on JF17K and WikiPeople-50bi. To see the reasons behind, we analyze the type information. We present the analyses on WikiPeople and other datasets get similar results. The visualization of the type embeddings of the role-values on WikiPeople is illustrated in Fig.~\ref{fig:type_embed_WikiPeople}, where the role-values of the same type are in the same label and the same color. Here, the widely used t-SNE~\cite{tSNE} is adopted to reduce the dimension of type embeddings. The statistics of the role-value types on WikiPeople (in logarithm) are illustrated in Fig.~\ref{fig:type_stat_WikiPeople}, where X-axis indicates the size of the role-value type ($n_{type}$), i.e., the number of the role-values in this type, and Y-axis indicates the number of the types of size $n_{type}$. Although the type embeddings of the role-values in tNaLP capture useful information (see Fig.~\ref{fig:type_embed_WikiPeople}), i.e., those of the same type are close to each other, as the type distribution is highly imbalanced (see Fig.~\ref{fig:type_stat_WikiPeople}), tNaLP does not demonstrate its superiority. Actually, this imbalance of type distribution on JF17K and WikiPeople-50bi is more serious. Thus, tNaLP performs much worse on JF17K and WikiPeople-50bi.

In a word, from Tables~\ref{table:extend_methods_comparison}, \ref{table:extend_methods_WikiPeople-n}, and \ref{table:extend_methods_WikiPeople-xbi}, we conclude that the new negative sampling mechanism is a good choice on the datasets of many n-ary relational facts, while introducing the type constraints is preferred when the type distribution of the datasets is not so imbalanced.

\subsection{Link Prediction Performance Comparison on N-ary Relational Data}
\subsubsection{Baselines}
\label{subsubsec:baselines}

\begin{table*}[htb]
	\setlength{\tabcolsep}{0.7em}
	\centering
	\caption{Experimental results of role prediction in terms of MRR and Hits@\{1, 3, 10\}.}
	\vspace{-1mm}
	\vspace{-1mm}
	\makebox[\columnwidth][c]{
		\begin{tabular}{c|ccc|ccc|ccc|ccc}
			\toprule[1pt]
			\multirow{2}{*}{Dataset} &\multicolumn{3}{c|}{MRR} &\multicolumn{3}{c|}{Hits@1} &\multicolumn{3}{c|}{Hits@3} &\multicolumn{3}{c}{Hits@10}\\
			\cline{2-13} &Binary &N-ary &Overall &Binary &N-ary &Overall &Binary &N-ary &Overall &Binary &N-ary &Overall\\
			\midrule[0.5pt]
			JF17K &0.940 &0.964 &0.956 &0.913 &0.949 &0.938 &0.961 &0.976 &0.971 &0.984 &0.988 &0.986\\ 
			WikiPeople &0.788 &0.795 &0.790 &0.712 &0.717 &0.713 &0.840 &0.855 &0.843 &0.927 &0.930 &0.927\\
			\bottomrule[1pt]
		\end{tabular}
	}
	\vspace{-2mm}
	\label{table:role_prediction}
\end{table*}

\begin{table*}[htb]
	\setlength{\tabcolsep}{0.97em}
	\centering
	\caption{Experimental results of role-value prediction in terms of MRR and Hits@\{1, 3, 10\} on WikiPeople.}
	\vspace{-1mm}
	\vspace{-1mm}
	\makebox[\columnwidth][c]{
		\begin{tabular}{c|ccc|ccc|ccc|ccc}
			\toprule[1pt]
			\multirow{2}{*}{Dataset} &\multicolumn{3}{c|}{MRR} &\multicolumn{3}{c|}{Hits@1} &\multicolumn{3}{c|}{Hits@3} &\multicolumn{3}{c}{Hits@10}\\
			\cline{2-13} &Binary &N-ary &Overall &Binary &N-ary &Overall &Binary &N-ary &Overall &Binary &N-ary &Overall\\
			\midrule[0.5pt]
			RAE &0.169 &0.187 &0.172 &0.096 &0.126 &0.102 &0.178 &0.198 &0.182 &0.323 &0.306 &0.320\\
			HypE &0.281 &0.286 &0.282 &0.138 &0.191 &0.148 &0.384 &0.320 &0.372 &\textbf{0.489} &0.478 &\textbf{0.487}\\
			\midrule[0.5pt]
			tNaLP &\textbf{0.366} &0.283 &\textbf{0.350} &\textbf{0.309} &0.193 &\textbf{0.288} &\textbf{0.390} &0.319 &\textbf{0.377} &0.473 &0.462 &0.471\\
			tNaLP\textsuperscript{+} &0.346 &\textbf{0.308} &0.339 &0.283 &\textbf{0.206} &0.269 &0.373 &\textbf{0.348} &0.369 &0.465 &\textbf{0.508} &0.473\\
			\bottomrule[1pt]
		\end{tabular}
	}
	\vspace{-2mm}
	\label{table:value_prediction}
\end{table*}

\begin{table*}[htb]
	\setlength{\tabcolsep}{0.7em}
	\centering
	\caption{Detailed experimental results of role-value prediction on each arity ($>$2) in terms of MRR and Hits@\{1, 3, 10\} on WikiPeople.}
	\vspace{-1mm}
	\vspace{-1mm}
	\makebox[\columnwidth][c]{
		\begin{tabular}{c|cccc|cccc|cccc|cccc}
			\toprule[1pt]
			\multirow{2}{*}{\!Method\!} &\multicolumn{4}{c|}{MRR} &\multicolumn{4}{c|}{Hits@1} &\multicolumn{4}{c|}{Hits@3} &\multicolumn{4}{c}{Hits@10}\\
			\cline{2-17} &3-ary &4-ary &5-ary &6-ary &3-ary &4-ary &5-ary &6-ary &3-ary &4-ary &5-ary &6-ary &3-ary &4-ary &5-ary &6-ary\\
			\midrule[0.5pt]
			HypE &0.266 &0.304 &0.370 &0.122 &0.183 &0.191 &\textbf{0.282} &0.058 &0.284 &0.360 &0.416 &0.118 &0.443 &0.527 &0.526 &0.282\\
			\midrule[0.5pt]
			tNaLP &0.247 &0.318 &0.367 &0.205 &0.170 &0.211 &0.264 &0.121 &0.267 &0.370 &0.430 &0.245 &0.403 &0.527 &0.549 &0.376\\
			tNaLP\textbf{+} &\textbf{0.270} &\textbf{0.344} &\textbf{0.387} &\textbf{0.250} &\textbf{0.185} &\textbf{0.223} &0.274 &\textbf{0.148} &\textbf{0.292} &\textbf{0.406} &\textbf{0.456} &\textbf{0.273} &\textbf{0.444} &\textbf{0.578} &\textbf{0.593} &\textbf{0.476}\\
			\bottomrule[1pt]
		\end{tabular}
	}
	\vspace{-4mm}
	\label{table:detailed_value_prediction}
\end{table*}

The representative or state-of-the-art methods for link prediction on n-ary relational data directly are RAE~\cite{RAE}, HypE~\cite{HypE}, GETD~\cite{GETD}, HINGE~\cite{HINGE}, and NeuInfer~\cite{NeuInfer}. As the tensor based method GETD inherently requires all facts to have the same arities, it is not applicable to JF17K and WikiPeople with mixed arities. To be fair, HINGE and NeuInfer are not adopted as baselines, since they introduce additional principal and subordinate structure information. Therefore, we adopt RAE and HypE as baselines.

\subsubsection{Experimental results}
\label{subsubsec:NaLP_SKG_results}
Link prediction on n-ary relational data has two tasks, i.e., role prediction and role-value prediction. 

As RAE and HypE are deliberately designed only for role-value prediction, we conduct role prediction only on the proposed tNaLP\textsuperscript{+} method. To elaborate the performance, we group the test set into binary and n-ary (n$>$2) categories according to the arities of the facts. Table~\ref{table:role_prediction} demonstrates the experimental results in terms of all the four metrics on binary category, n-ary category, and the whole dataset. The experimental results illustrate the power of tNaLP\textsuperscript{+}. It achieves high performance on all the metrics. We attribute this to the reasonable modeling of r:v pairs.

For role-value prediction, we select tNaLP and tNaLP\textsuperscript{+} to compare with, based on the comparison results in Table~\ref{table:extend_methods_comparison}, and conduct experiments on larger and more difficult WikiPeople for space limitation. The experimental results of role-value prediction compared to RAE and HypE are reported in Table~\ref{table:value_prediction}.

From the results, it is clear that tNaLP and tNaLP\textsuperscript{+} perform better than the baselines. It testifies the strength of considering the relatedness of r:v pairs, introducing the type constraints, and proposing the more reasonable negative sampling mechanism. The tNaLP and tNaLP\textsuperscript{+} methods even improve the overall performance by 14.0\% and 12.1\% on Hits@1, respectively, compared to the best baseline HypE, which elucidates that the proposed methods are more effective in picking exactly valid role-values out. We observe that RAE performs much worse than other methods. It is not surprising, since tNaLP and tNaLP\textsuperscript{+} are capable of better coping with diverse data. Data incompleteness/insertion/update, which is ubiquitous on WikiPeople, is handled elegantly in tNaLP and tNaLP\textsuperscript{+}, while new relation for it is defined in RAE and HypE, which may lead to data sparsity when the instances of the new relation are in small number. Although position-specific convolution further improves the performance of HypE, it still underperforms the proposed methods.

To further comprehensively evaluate the performance of tNaLP and tNaLP\textsuperscript{+} on role-value prediction, we select the best baseline from Table~\ref{table:value_prediction}, i.e., HypE, and present the detailed experimental results on each arity ($>$2) in Table~\ref{table:detailed_value_prediction} (the arities having less than 10 facts are not reported). It can be seen from Table~\ref{table:detailed_value_prediction} that the proposed methods demonstrate their advantage on each arity.

From Tables~\ref{table:value_prediction} and \ref{table:detailed_value_prediction}, we observe that tNaLP\textsuperscript{+} performs better than tNaLP on n-ary relational facts, but worse on binary relational facts. It testified our conclusion in Section~\ref{subsec:NaLP_extend_effectiveness} that the new negative sampling mechanism benefits n-ary relational facts.

\begin{figure*}[!htb]
	\centering
	\includegraphics[width=5.0in]{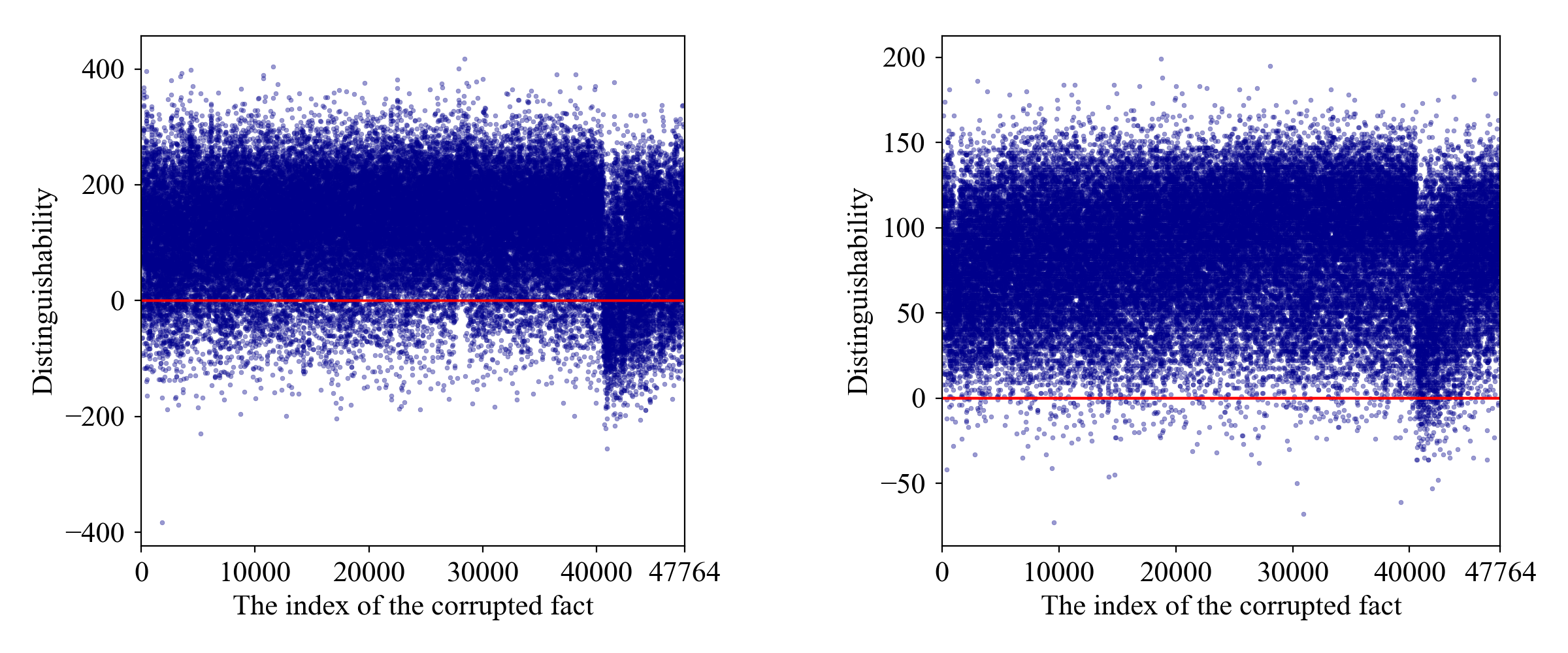}
	\vspace{-3mm}
	\vspace{-2mm}
	\caption{The visualization of the distinguishability results of the binary Case-1 (left) and the n-ary Case-2 (right) on NaLP.}
	\label{fig:case_study}
	\vspace{-1mm}
\end{figure*}

\begin{table*}[htb]
	\centering
	\caption{The replaced role-values in the top 10 corrupted facts of Case-2 (awards are marked in boldface, and the number of awards in the top 10 list on each method is presented in the second row).}
	\vspace{-1mm}
	\vspace{-1mm}
	\makebox[\columnwidth][c]{
		\begin{tabular}{c|c|c|c}
			\toprule[1pt]
			NaLP &tNaLP &NaLP\textsuperscript{+} &tNaLP\textsuperscript{+}\\
			\midrule[0.5pt]
			0 &5 &3 &6\\
			\midrule[0.5pt]
			Marie Curie  &Marie Curie &\textbf{Poncelet Prize} &\textbf{75th Academy Awards}\\
			\midrule[0.5pt]
			Henri Becquerel &\textbf{Rumford Medal} &2018 &Aimé Cotton\\
			\midrule[0.5pt]
			1903 &1903 &2008 &\textbf{Nobel Prize in Chemistry}\\
			\midrule[0.5pt]
			1893 &Rodion Malinovsky &FK AS Trenčín &\textbf{Lilienfeld Prize}\\
			\midrule[0.5pt]
			Auguste Perret &\textbf{75th Academy Awards} &\textbf{Paleontological Society Medal} &Santos F.C.\\
			\midrule[0.5pt]
			R.C. Lens &FK AS Trenčín &Sociedade Esportiva Palmeiras &1918\\
			\midrule[0.5pt]
			Aílton Gonçalves da Silva &\textbf{Franklin Medal} &1933 &\textbf{ASME Medal}\\
			\midrule[0.5pt]
			Spanish Royal Academy of Sciences &\textbf{Wolf Prize in Physics} &\textbf{Henry Draper Medal} &UEFA Euro 1996\\
			\midrule[0.5pt]
			\makecell[c]{Officer's Cross of the Order of Merit\\ of the Federal Republic of Germany} &Glendale &Club América &\textbf{Hanns Martin Schleyer Prize}\\
			\midrule[0.5pt]
			Percy Williams Bridgman &\makecell[c]{\textbf{Centenary Prize der} \\\textbf{Royal Society of Chemistry}} &Nagoya Grampus &\textbf{Howard N. Potts Medal}\\
			\bottomrule[1pt]
		\end{tabular}
	}
	\vspace{-4mm}
	\label{table:top_corrupted_samples}
\end{table*}

\subsection{Overall Relatedness Analysis}
In the proposed methods, the overall relatedness feature vector of a set of r:v pairs, i.e., an n-ary relational fact, is the crucial intermediate result. We conduct further analyses to dig deep into what it has learned. 

\subsubsection{Distinguishability Metric}
According to Section~\ref{subsec:NaLP_framework}, a valid n-ary relational fact is expected to have an overall relatedness feature vector of large values, while an invalid n-ary relational fact is on the contrary. How to evaluate the degree of largeness? Actually, the relative magnitude between a valid n-ary relational fact and its corrupted facts is relatively more meaningful. Specifically, it is expected that a valid n-ary relational fact has larger values in a majority of dimensions of the overall relatedness feature vector compared to its corrupted facts, and thus the valid n-ary relational fact is distinguishable from its corrupted facts. Therefore, we propose the following metric to measure this type of distinguishability:
\begin{equation}
\begin{split}
d(Rel\textsuperscript{+}, Rel\textsuperscript{-}) =\,\, &{\rm sum}\left({\rm sgn}(\mathbf{\mathcal{R}_{Rel\textsuperscript{+}}} - \mathbf{\mathcal{R}_{Rel\textsuperscript{-}}})\right)-\\
&{\rm sum}\left({\rm sgn}(\mathbf{\mathcal{R}_{Rel\textsuperscript{-}}} - \mathbf{\mathcal{R}_{Rel\textsuperscript{+}}})\right),
\label{eq:distinguishability}	
\end{split}
\end{equation}
where $Rel\textsuperscript{+}$ is a valid n-ary relational fact, and $Rel\textsuperscript{-}$ is one of its corrupted facts; $\mathbf{\mathcal{R}_{Rel\textsuperscript{+}}}$ and $\mathbf{\mathcal{R}_{Rel\textsuperscript{-}}}$ are the overall relatedness feature vectors of $Rel\textsuperscript{+}$ and $Rel\textsuperscript{-}$, respectively; ${\rm sgn}(x)$ is the function that returns 1, if $x>0$, otherwise, returns 0; ${\rm sum}(\cdot)$ is the element-wise sum function. In Equation~\eqref{eq:distinguishability}, the left part of the minus sign counts the number of dimensions that $\mathbf{\mathcal{R}_{Rel\textsuperscript{+}}}$ is larger than $\mathbf{\mathcal{R}_{Rel\textsuperscript{-}}}$, and the right part is defined similarly. Hence, $d(Rel\textsuperscript{+}, Rel\textsuperscript{-})$ measures the relative amount that $\mathbf{\mathcal{R}_{Rel\textsuperscript{+}}}$ has more dimensions of larger values than $\mathbf{\mathcal{R}_{Rel\textsuperscript{-}}}$.

\subsubsection{Case Study}
Based on this proposed distinguishability metric, we select two typical cases from the validation set of WikiPeople, one binary case and one n-ary case, which are predicted correctly by NaLP, to carry out overall relatedness analysis. Note that, we only sample cases regarding more difficult role-value prediction. The sampled cases are as follows:
\begin{itemize}
	\item Case-1: Predict $Michael\ Douglas$, given the role $son$ and the remaining r:v pair of the binary relational fact $\{son:Michael$ $Douglas,$ $father:$ $Kirk$ $Douglas\}$, denoted as Fact-1.
	\item Case-2: Predict $Nobel\ Prize\ in\ Physics$, given the role $award$ and the remaining r:v pairs of the n-ary relational fact $\{person:Marie$ $Curie,$ $award:Nobel$ $Prize$ $in$ $Physics,$ $point$ $in$ $time\!:$ $\textit{1903},$ $together$ $with\!:$ $Henri$ $Becquerel,$ $together$ $with:$ $Pierre\ Curie\}$, denoted as Fact-2.
\end{itemize}
Similar to the evaluation procedure, we replace $Michael$ $Douglas$ in Fact-1 and $Nobel\ Prize\ in\ Physics$ in Fact-2 with all the other role-values in $V$, then these corrupted facts, as well as Fact-1 and Fact-2 are fed into the proposed methods to obtain the overall relatedness feature vectors. Subsequently, the distinguishability metric is computed via Equation~\eqref{eq:distinguishability}. Fig.~\ref{fig:case_study} depicts these distinguishability results on NaLP and the extended methods get similar results.

As exhibited in Fig.~\ref{fig:case_study}, most distinguishability results lie in the area above 0. It visually corroborates that the overall relatedness feature vector of a fact captures many discriminant features to further estimate its validity. Therefore, our conjecture on relatedness (see Section~\ref{subsec:NaLP_framework}) makes sense to a certain degree. Due to the diversity of n-ary relational facts, the dimension of the overall relatedness feature vectors is encouraged to be large. In this way, there are sufficient dimensions to encode various features. This is consistent with the large optimal value of $n_{\text{gFCN}}$ (e.g., $800$ on JF17K and $1200$ on WikiPeople).

There are also some corrupted facts that obtain extremely small distinguishability results and it is difficult to distinguish them from the valid facts. We take Case-2 for example and present the replaced role-values in its top 10 corrupted facts on NaLP and the extended methods in Table~\ref{table:top_corrupted_samples}.

\begin{figure*}[!htb]
	\centering
	\includegraphics[width=6in]{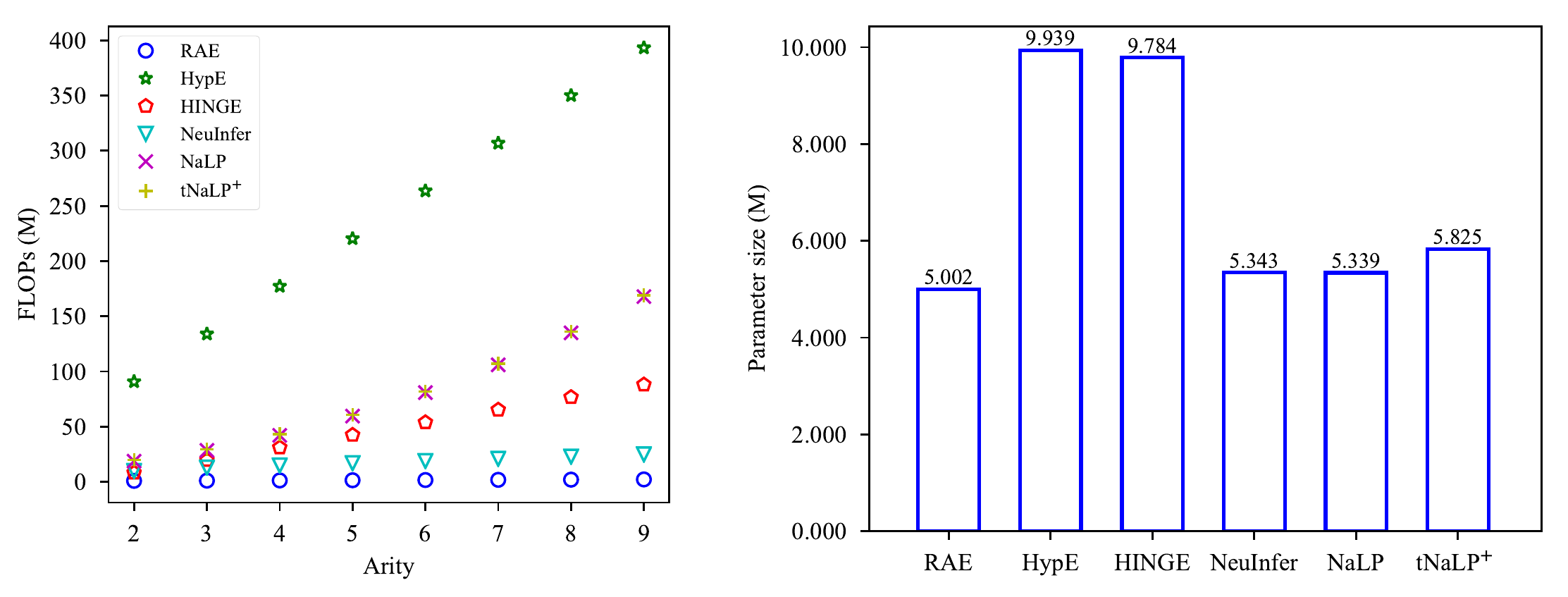}
	\vspace{-3mm}
	\vspace{-2mm}
	\caption{Complexity comparison on WikiPeople.}
	\label{fig:complexity_analysis}
	\vspace{-1mm}
	\vspace{-1mm}
	\vspace{-1mm}
	\vspace{-3mm}
\end{figure*}

From Table~\ref{table:top_corrupted_samples}, we can observe that on NaLP, the corrupted facts with the replaced role-values participating in some other roles of Fact-2, i.e, $Marie$ $Curie$, $Henri$ $Becquerel$, and $\textit{1903}$, obtain the top 3 smallest distinguishability results, respectively. NaLP is unable to filter out such a situation. Some corrupted facts with the replaced role-values of type $time$, $club$, or $institution$ also achieve relatively smaller distinguishability results. It is acceptable, since NaLP does not take role-value type into consideration. By introducing type constraints or the new negative sampling mechanism, tNaLP and NaLP\textsuperscript{+} learn more reasonable embeddings, and some corrupted facts with the replaced role-values of type $award$ get the smallest distinguishability results. By combining type constraints and the new negative sampling mechanism, tNaLP\textsuperscript{+} further makes more corrupted facts with the replaced role-values of type $award$ (6 out of the top 10 corrupted facts) get the smallest distinguishability results. These role-values are of the same type as the right role-value $Nobel\ Prize\ in\ Physics$ and are difficult to filter out, especially for $Wolf\ Prize\ in\ Physics$, $Centenary\ Prize\ der\ Royal\ Society\ of\ Chemistry$, $Nobel\ Prize\ in\ Chemistry$, and $Lilienfeld\ Prize$ (an award administered by the American Physical Society), since $Marie$ $Curie$ is a physicist and chemist, and the right role-value $Nobel\ Prize\ in\ Physics$ is a Nobel Prize. Thus, it demonstrates that the extended methods tNaLP, NaLP\textsuperscript{+}, and tNaLP\textsuperscript{+} are able to learn relatively more useful information.

\subsection{Ablation Study}
To look deep into the reasonability of the design choices in Section~\ref{sec:NaLP_method}, we perform an ablation study on NaLP. Without loss of generality, these experiments are conducted on WikiPeople concerning more difficult role-value prediction. Detailedly, convolution in the r:v pair embedding component is replaced by plus and multiplication, denoted as NaLP(plus) and NaLP(mul), respectively. Besides, element-wise minimizing in the relatedness evaluation component is replaced by element-wise maximizing and mean, denoted as NaLP(emax) and NaLP(emean), respectively. The experimental comparison between NaLP and the ablation methods is presented in Table~\ref{table:ablation_study}. It can be observed from the table that NaLP outperforms all the ablation methods significantly. It suggests that convolution is a much better way to learn features than plus and multiplication. Moreover, element-wise minimizing is experimentally more reasonable than element-wise maximizing and mean.
\begin{table}[!htb]
\vspace{-3mm}
	\setlength{\tabcolsep}{1em}
	\centering
	\caption{The experimental comparison of role-value prediction between NaLP and the ablation methods on WikiPeople.}
	\vspace{-1mm}
	\vspace{-2mm}
	\makebox[\columnwidth][c]{
		\begin{tabular}{c|cccc}
			\toprule[1pt]
			Method &MRR &Hits@1 &Hits@3 &Hits@10\\
			\midrule[0.5pt]
			NaLP &\textbf{0.338} &\textbf{0.272} &\textbf{0.364} &\textbf{0.466}\\
			NaLP(plus) &0.156 &0.107 &0.188 &0.227\\
			NaLP(mul) &0.305 &0.242 &0.331 &0.420\\
			NaLP(emax) &0.312 &0.243 &0.338 &0.443\\
			NaLP(emean) &0.222 &0.147 &0.255 &0.359\\
			\bottomrule[1pt]
		\end{tabular}
	}
	\vspace{-5mm}
	\label{table:ablation_study}
\end{table}

\subsection{Complexity Analysis}
\label{subsec:complexity_analysis}
To see the efficiency of the proposed methods, we compare them with the representative or state-of-the-art link prediction methods on n-ary relational data, in terms of FLOPs (floating point of operations) and parameter size on the larger WikiPeople. The comparison is illustrated in Fig.~\ref{fig:complexity_analysis}. As in GETD~\cite{GETD}, all facts are required to have the same arities, which is not reasonable, thus we do not compare with it. It can be observed from the figure that NaLP and tNaLP\textsuperscript{+} have similar FLOPs. Although they have more FLOPs than RAE~\cite{RAE}, HINGE~\cite{HINGE}, and NeuInfer~\cite{NeuInfer}, they are much less time-consuming them HypE~\cite{HypE}. As for parameter size, all the methods have more than 5.000M parameters. The proposed methods are comparable with NeuInfer and have much less parameters than HypE and HINGE.

\section{Conclusions and Future Works}
\label{sec:conclusion_futurework}
In this paper, we represented each n-ary relational fact as a set of its r:v pairs and proposed NaLP and its extensions to cope with link prediction on n-ary relational data. The design of the frameworks equips the proposed methods with the characteristic of permutation invariance to the input order of the r:v pairs and the ability to well handle facts of different arities. By evaluating the relatedness of every two r:v pairs in an n-ary relational fact, NaLP is able to estimate the overall relatedness of all its r:v pairs approximately. The resulting overall relatedness feature vectors further enable NaLP to pick up many determinant features to decide the validity of the input facts. We further extended NaLP to tNaLP, NaLP\textsuperscript{+}, and tNaLP\textsuperscript{+} via introducing type constraints of roles and role-values unsupervisedly, a more reasonable negative sampling mechanism, and both, respectively. Furthermore, since publicly available n-ary relational datasets are limited, we developed a practical one, WikiPeople, and further derived WikiPeople-n, WikiPeople-0bi, WikiPeople-50bi, and WikiPeople-100bi from WikiPeople, with different percentages of binary relational facts. They were published online for further research. Experimental results on the public dataset JF17K and the newly developed datasets manifest the merits and superiority of the proposed methods. Specifically, on role-value prediction, compared to the state-of-the-art method, tNaLP and tNaLP\textsuperscript{+} improve the performance even by 14.0\% and 12.1\% in terms of Hits@1, respectively.

For future works, on the one hand, we will explore more expressive neural network models to capture more favorable features for relatedness evaluation. On the other hand, in this paper, we use only facts in the datasets to conduct link prediction on n-ary relational data. In the future, we will introduce additional information, such as rules and external texts, to further improve the developed methods. Moreover, type constraints of roles and role-values without supervision and the more reasonable negative sampling mechanism can be adopted to promote other state-of-the-art link prediction methods on n-ary relational data, such as the newly published NeuInfer~\cite{NeuInfer}.

\ifCLASSOPTIONcompsoc
  \section*{Acknowledgments}
\else
  \section*{Acknowledgment}
\fi

The work was supported in part by the National Key Research and Development Program of China under Grant 2016YFB1000902, in part by the Beijing Academy of Artificial Intelligence (BAAI) under Grant BAAI2019ZD0306, in part by the Lenovo-CAS Joint Lab Youth Scientist Project, in part by the National Natural Science Foundation of China under Grants 62002341, U1911401, 61772501, U1836206, and 91646120, and in part by the GFKJ Innovation Program.

\ifCLASSOPTIONcaptionsoff
  \newpage
\fi

\bibliographystyle{IEEEtran}
\bibliography{TKDE20_extend}

\begin{thebibliography}{10}
\providecommand{\url}[1]{#1}
\csname url@samestyle\endcsname
\providecommand{\newblock}{\relax}
\providecommand{\bibinfo}[2]{#2}
\providecommand{\BIBentrySTDinterwordspacing}{\spaceskip=0pt\relax}
\providecommand{\BIBentryALTinterwordstretchfactor}{4}
\providecommand{\BIBentryALTinterwordspacing}{\spaceskip=\fontdimen2\font plus
\BIBentryALTinterwordstretchfactor\fontdimen3\font minus
  \fontdimen4\font\relax}
\providecommand{\BIBforeignlanguage}[2]{{%
\expandafter\ifx\csname l@#1\endcsname\relax
\typeout{** WARNING: IEEEtran.bst: No hyphenation pattern has been}%
\typeout{** loaded for the language `#1'. Using the pattern for}%
\typeout{** the default language instead.}%
\else
\language=\csname l@#1\endcsname
\fi
#2}}
\providecommand{\BIBdecl}{\relax}
\BIBdecl

\bibitem{Freebase}
K.~Bollacker, C.~Evans, P.~Paritosh, T.~Sturge, and J.~Taylor, ``Freebase: A
  collaboratively created graph database for structuring human knowledge,'' in
  \emph{Proceedings of the 2008 ACM SIGMOD International Conference on
  Management of Data}, 2008, pp. 1247--1250.

\bibitem{m-TransH}
J.~Wen, J.~Li, Y.~Mao, S.~Chen, and R.~Zhang, ``On the representation and
  embedding of knowledge bases beyond binary relations,'' in \emph{Proceedings
  of the 25th International Joint Conference on Artificial Intelligence}, 2016,
  pp. 1300--1307.

\bibitem{nickel2016review}
M.~Nickel, K.~Murphy, V.~Tresp, and E.~Gabrilovich, ``A review of relational
  machine learning for knowledge graphs,'' \emph{Proceedings of the IEEE}, vol.
  104, no.~1, pp. 11--33, 2016.

\bibitem{wang2017review}
Q.~Wang, Z.~Mao, B.~Wang, and L.~Guo, ``Knowledge graph embedding: A survey of
  approaches and applications,'' \emph{IEEE Transactions on Knowledge and Data
  Engineering}, vol.~29, no.~12, pp. 2724--2743, 2017.

\bibitem{GE_survey}
H.~Cai, V.~W. Zheng, and K.~C.-C. Chang, ``A comprehensive survey of graph
  embedding: Problems, techniques, and applications,'' \emph{IEEE Transactions
  on Knowledge and Data Engineering}, vol.~30, no.~9, pp. 1616--1637, 2018.

\bibitem{RAE}
R.~Zhang, J.~Li, J.~Mei, and Y.~Mao, ``Scalable instance reconstruction in
  knowledge bases via relatedness affiliated embedding,'' in \emph{Proceedings
  of the 27th International Conference on World Wide Web}, 2018, pp.
  1185--1194.

\bibitem{HypE}
B.~Fatemi, P.~Taslakian, D.~Vázquez, and D.~Poole, ``Knowledge hypergraphs:
  Extending knowledge graphs beyond binary relations,'' in \emph{Proceedings of
  the 29th International Joint Conference on Artificial Intelligence}, 2020,
  pp. 2191--2197.

\bibitem{GETD}
Y.~Liu, Q.~Yao, and Y.~Li, ``Generalizing tensor decomposition for n-ary
  relational knowledge bases,'' in \emph{Proceedings of the 29th International
  Conference on World Wide Web}, 2020, pp. 1104--1114.

\bibitem{HINGE}
P.~Rosso, D.~Yang, and P.~Cudr\'{e}-Mauroux, ``Beyond triplets:
  Hyper-relational knowledge graph embedding for link prediction,'' in
  \emph{Proceedings of the 29th International Conference on World Wide Web},
  2020, pp. 1885--1896.

\bibitem{NeuInfer}
S.~Guan, X.~Jin, J.~Guo, Y.~Wang, and X.~Cheng, ``Neuinfer: Knowledge inference
  on n-ary facts,'' in \emph{Proceedings of the 58th Annual Meeting of the
  Association for Computational Linguistics}, 2020, pp. 6141--6151.

\bibitem{Wikidata}
D.~Vrande{\v{c}}i{\'c} and M.~Kr{\"o}tzsch, ``Wikidata: A free collaborative
  knowledgebase,'' \emph{Communications of the ACM}, vol.~57, no.~10, pp.
  78--85, 2014.

\bibitem{RESCAL}
M.~Nickel, V.~Tresp, and H.-P. Kriegel, ``A three-way model for collective
  learning on multi-relational data,'' in \emph{Proceedings of the 28th
  International Conference on Machine Learning}, 2011, pp. 809--816.

\bibitem{nickel2014reducing}
M.~Nickel, X.~Jiang, and V.~Tresp, ``Reducing the rank in relational
  factorization models by including observable patterns,'' in \emph{Proceedings
  of the 27th International Conference on Neural Information Processing
  Systems}, 2014, pp. 1179--1187.

\bibitem{krompabeta2015type}
D.~Krompa$\beta$, S.~Baier, and V.~Tresp, ``Type-constrained representation
  learning in knowledge graphs,'' in \emph{Proceedings of the 14th
  International Conference on The Semantic Web}, 2015, pp. 640--655.

\bibitem{ComplEx}
T.~Trouillon, J.~Welbl, S.~Riedel, E.~Gaussier, and G.~Bouchard, ``Complex
  embeddings for simple link prediction,'' in \emph{Proceedings of the 33rd
  International Conference on Machine Learning}, vol.~48, 2016, pp. 2071--2080.

\bibitem{RSTE}
Y.~Tay, A.~T. Luu, S.~C. Hui, and F.~Brauer, ``Random semantic tensor ensemble
  for scalable knowledge graph link prediction,'' in \emph{Proceedings of the
  10th ACM International Conference on Web Search and Data Mining}, 2017, pp.
  751--760.

\bibitem{ComplEx_modify}
H.~Manabe, K.~Hayashi, and M.~Shimbo, ``Data-dependent learning of
  symmetric/antisymmetric relations for knowledge base completion,'' in
  \emph{Proceedings of the 32nd AAAI Conference on Artificial Intelligence},
  2018, pp. 3754--3761.

\bibitem{ComplEx_modify_cn}
B.~Ding, Q.~Wang, B.~Wang, and L.~Guo, ``Improving knowledge graph embedding
  using simple constraints,'' in \emph{Proceedings of the 56th Annual Meeting
  of the Association for Computational Linguistics}, 2018, pp. 110--121.

\bibitem{TypeComplex}
P.~Jain, P.~Kumar, S.~Chakrabarti \emph{et~al.}, ``Type-sensitive knowledge
  base inference without explicit type supervision,'' in \emph{Proceedings of
  the 56th Annual Meeting of the Association for Computational Linguistics
  (Volume 2: Short Papers)}.\hskip 1em plus 0.5em minus 0.4em\relax Association
  for Computational Linguistics, 2018, pp. 75--80.

\bibitem{TransE}
A.~Bordes, N.~Usunier, A.~Garcia-Duran, J.~Weston, and O.~Yakhnenko,
  ``Translating embeddings for modeling multi-relational data,'' in
  \emph{Proceedings of the 26th International Conference on Neural Information
  Processing Systems}, 2013, pp. 2787--2795.

\bibitem{TransH}
Z.~Wang, J.~Zhang, J.~Feng, and Z.~Chen, ``Knowledge graph embedding by
  translating on hyperplanes,'' in \emph{Proceedings of the 28th AAAI
  Conference on Artificial Intelligence}, 2014, pp. 1112--1119.

\bibitem{TransR}
Y.~Lin, Z.~Liu, M.~Sun, Y.~Liu, and X.~Zhu, ``Learning entity and relation
  embeddings for knowledge graph completion,'' in \emph{Proceedings of the 29th
  AAAI Conference on Artificial Intelligence}, 2015, pp. 2181--2187.

\bibitem{TransD}
G.~Ji, S.~He, L.~Xu, K.~Liu, and J.~Zhao, ``Knowledge graph embedding via
  dynamic mapping matrix,'' in \emph{Proceedings of the 53rd Annual Meeting of
  the Association for Computational Linguistics}, 2015, pp. 687--696.

\bibitem{PTransE}
Y.~Lin, Z.~Liu, H.~Luan, M.~Sun, S.~Rao, and S.~Liu, ``Modeling relation paths
  for representation learning of knowledge bases,'' in \emph{Proceedings of the
  2015 Conference on Empirical Methods in Natural Language Processing}, 2015,
  pp. 705--714.

\bibitem{TransG}
H.~Xiao, M.~Huang, and X.~Zhu, ``{TransG}: A generative model for knowledge
  graph embedding,'' in \emph{Proceedings of the 54th Annual Meeting of the
  Association for Computational Linguistics}, 2016, pp. 2316--2325.

\bibitem{TranSparse}
G.~Ji, K.~Liu, S.~He, and J.~Zhao, ``Knowledge graph completion with adaptive
  sparse transfer matrix,'' in \emph{Proceedings of the 30th AAAI Conference on
  Artificial Intelligence}, 2016, pp. 985--991.

\bibitem{TransA}
Y.~Jia, Y.~Wang, H.~Lin, X.~Jin, and X.~Cheng, ``Locally adaptive translation
  for knowledge graph embedding,'' in \emph{Proceedings of the 30th AAAI
  Conference on Artificial Intelligence}, 2016, pp. 992--998.

\bibitem{TKRL}
R.~Xie, Z.~Liu, and M.~Sun, ``Representation learning of knowledge graphs with
  hierarchical types,'' in \emph{Proceedings of the 25th International Joint
  Conference on Artificial Intelligence}, 2016, pp. 2965--2971.

\bibitem{TransT}
S.~Ma, J.~Ding, W.~Jia, K.~Wang, and M.~Guo, ``{TransT}: Type-based multiple
  embedding representations for knowledge graph completion,'' in \emph{Joint
  European Conference on Machine Learning and Knowledge Discovery in
  Databases}, 2017, pp. 717--733.

\bibitem{SSE}
S.~Guo, Q.~Wang, B.~Wang, L.~Wang, and L.~Guo, ``Semantically smooth knowledge
  graph embedding,'' in \emph{Proceedings of the 53rd Annual Meeting of the
  Association for Computational Linguistics (Volume 1: Long Papers)}, 2015, pp.
  84--94.

\bibitem{SSE_extend}
------, ``{SSE}: Semantically smooth embedding for knowledge graphs,''
  \emph{IEEE Transactions on Knowledge and Data Engineering}, vol.~29, no.~4,
  pp. 884--897, 2017.

\bibitem{TorusE}
T.~Ebisu and R.~Ichise, ``{TorusE}: Knowledge graph embedding on a {Lie}
  group,'' in \emph{Proceedings of the 32nd AAAI Conference on Artificial
  Intelligence}, 2018, pp. 1819--1826.

\bibitem{TorusE_extend}
------, ``Generalized translation-based embedding of knowledge graph,''
  \emph{IEEE Transactions on Knowledge and Data Engineering}, vol.~32, no.~5,
  pp. 941--951, 2020.

\bibitem{TAPR}
Y.~Shen, N.~Ding, H.~Zheng, Y.~Li, and M.~Yang, ``Modeling relation paths for
  knowledge graph completion,'' \emph{IEEE Transactions on Knowledge and Data
  Engineering}, 2020.

\bibitem{ER-MLP}
X.~Dong, E.~Gabrilovich, G.~Heitz, W.~Horn, N.~Lao, K.~Murphy, T.~Strohmann,
  S.~Sun, and W.~Zhang, ``Knowledge {Vault}: A web-scale approach to
  probabilistic knowledge fusion,'' in \emph{Proceedings of the 20th ACM SIGKDD
  International Conference on Knowledge Discovery and Data Mining}.\hskip 1em
  plus 0.5em minus 0.4em\relax ACM, 2014, pp. 601--610.

\bibitem{ProjE}
B.~Shi and T.~Weninger, ``{ProjE}: Embedding projection for knowledge graph
  completion,'' in \emph{Proceedings of the 31st AAAI Conference on Artificial
  Intelligence}, 2017, pp. 1236--1242.

\bibitem{R-GCN}
M.~Schlichtkrull, T.~N. Kipf, P.~Bloem, R.~v.~d. Berg, I.~Titov, and
  M.~Welling, ``Modeling relational data with graph convolutional networks,''
  in \emph{Proceedings of the 15th Extended Semantic Web Conference}, 2018, pp.
  593--607.

\bibitem{ConvE}
T.~Dettmers, P.~Minervini, P.~Stenetorp, and S.~Riedel, ``Convolutional {2D}
  knowledge graph embeddings,'' in \emph{Proceedings of the 32nd AAAI
  Conference on Artificial Intelligence}, 2018, pp. 1811--1818.

\bibitem{SENN}
S.~Guan, X.~Jin, Y.~Wang, and X.~Cheng, ``Shared embedding based neural
  networks for knowledge graph completion,'' in \emph{Proceedings of the 27th
  ACM International Conference on Information and Knowledge Management}, 2018,
  pp. 247--256.

\bibitem{ConvKB}
D.~Q. Nguyen, T.~D. Nguyen, D.~Q. Nguyen, and D.~Phung, ``A novel embedding
  model for knowledge base completion based on convolutional neural network,''
  in \emph{Proceedings of the 16th Annual Conference of the North American
  Chapter of the Association for Computational Linguistics: Human Language
  Technologies}, vol.~2, 2018, pp. 327--333.

\bibitem{SACN}
C.~Shang, Y.~Tang, J.~Huang, J.~Bi, X.~He, and B.~Zhou, ``End-to-end
  structure-aware convolutional networks for knowledge base completion,'' in
  \emph{Proceedings of the 33rd AAAI Conference on Artificial Intelligence},
  2019, pp. 3060--3067.

\bibitem{C-RNN}
Q.~Zhu, X.~Zhou, J.~Tan, and L.~Guo, ``Knowledge base reasoning with
  convolutional-based recurrent neural networks,'' \emph{IEEE Transactions on
  Knowledge and Data Engineering}, 2019.

\bibitem{CompGCN}
S.~Vashishth, S.~Sanyal, V.~Nitin, and P.~Talukdar, ``Composition-based
  multi-relational graph convolutional networks,'' in \emph{Proceedings of the
  8th International Conference on Learning Representations}, 2020, pp. 1--16.

\bibitem{DistMult}
B.~Yang, W.-t. Yih, X.~He, J.~Gao, and L.~Deng, ``Embedding entities and
  relations for learning and inference in knowledge bases,'' in
  \emph{Proceedings of the 3rd International Conference on Learning
  Representations}, 2015, pp. 1--13.

\bibitem{Permutate_eq}
N.~Guttenberg, N.~Virgo, O.~Witkowski, H.~Aoki, and R.~Kanai,
  ``Permutation-equivariant neural networks applied to dynamics prediction,''
  \emph{arXiv preprint arXiv:1612.04530}, 2016.

\bibitem{Deep_sets}
M.~Zaheer, S.~Kottur, S.~Ravanbhakhsh, B.~P\'{o}czos, R.~Salakhutdinov, and
  A.~J. Smola, ``Deep sets,'' in \emph{Proceedings of the 31st International
  Conference on Neural Information Processing Systems}, 2017, p. 3394–3404.

\bibitem{RNs}
A.~Santoro, D.~Raposo, D.~G. Barrett, M.~Malinowski, R.~Pascanu, P.~Battaglia,
  and T.~Lillicrap, ``A simple neural network module for relational
  reasoning,'' in \emph{Proceedings of the 31st Annual Conference on Neural
  Information Processing Systems}, 2017, pp. 4967--4976.

\bibitem{ReLU}
V.~Nair and G.~E. Hinton, ``Rectified linear units improve restricted boltzmann
  machines,'' in \emph{Proceedings of the 27th International Conference on
  International Conference on Machine Learning}, 2010, pp. 807--814.

\bibitem{Adam}
D.~Kingma and J.~Ba, ``Adam: A method for stochastic optimization,''
  \emph{arXiv preprint arXiv:1412.6980}, 2014.

\bibitem{Guu_traverseKG}
K.~Guu, J.~Miller, and P.~Liang, ``Traversing knowledge graphs in vector
  space,'' in \emph{Proceedings of the 2015 Conference on Empirical Methods in
  Natural Language Processing}, 2015, pp. 318--327.

\bibitem{xavier_initializer}
X.~Glorot and Y.~Bengio, ``Understanding the difficulty of training deep
  feedforward neural networks,'' in \emph{Proceedings of the 13rd International
  Conference on Artificial Intelligence and Statistics}, 2010, pp. 249--256.

\bibitem{BN}
S.~Ioffe and C.~Szegedy, ``Batch normalization: Accelerating deep network
  training by reducing internal covariate shift,'' in \emph{Proceedings of the
  32nd International Conference on Machine Learning}, 2015, pp. 448--456.

\bibitem{HolE}
M.~Nickel, L.~Rosasco, and T.~Poggio, ``Holographic embeddings of knowledge
  graphs,'' in \emph{Proceedings of the 30th AAAI Conference on Artificial
  Intelligence}, 2016, pp. 1955--1961.

\bibitem{tSNE}
L.~v.~d. Maaten and G.~Hinton, ``Visualizing data using {t-SNE},''
  \emph{Journal of machine learning research}, vol.~9, no. Nov, pp. 2579--2605,
  2008.

\end{thebibliography}

%
%

%
\vspace{-15mm}
\begin{IEEEbiography}
	[{\includegraphics[width=1in,height=1.25in,clip,keepaspectratio]{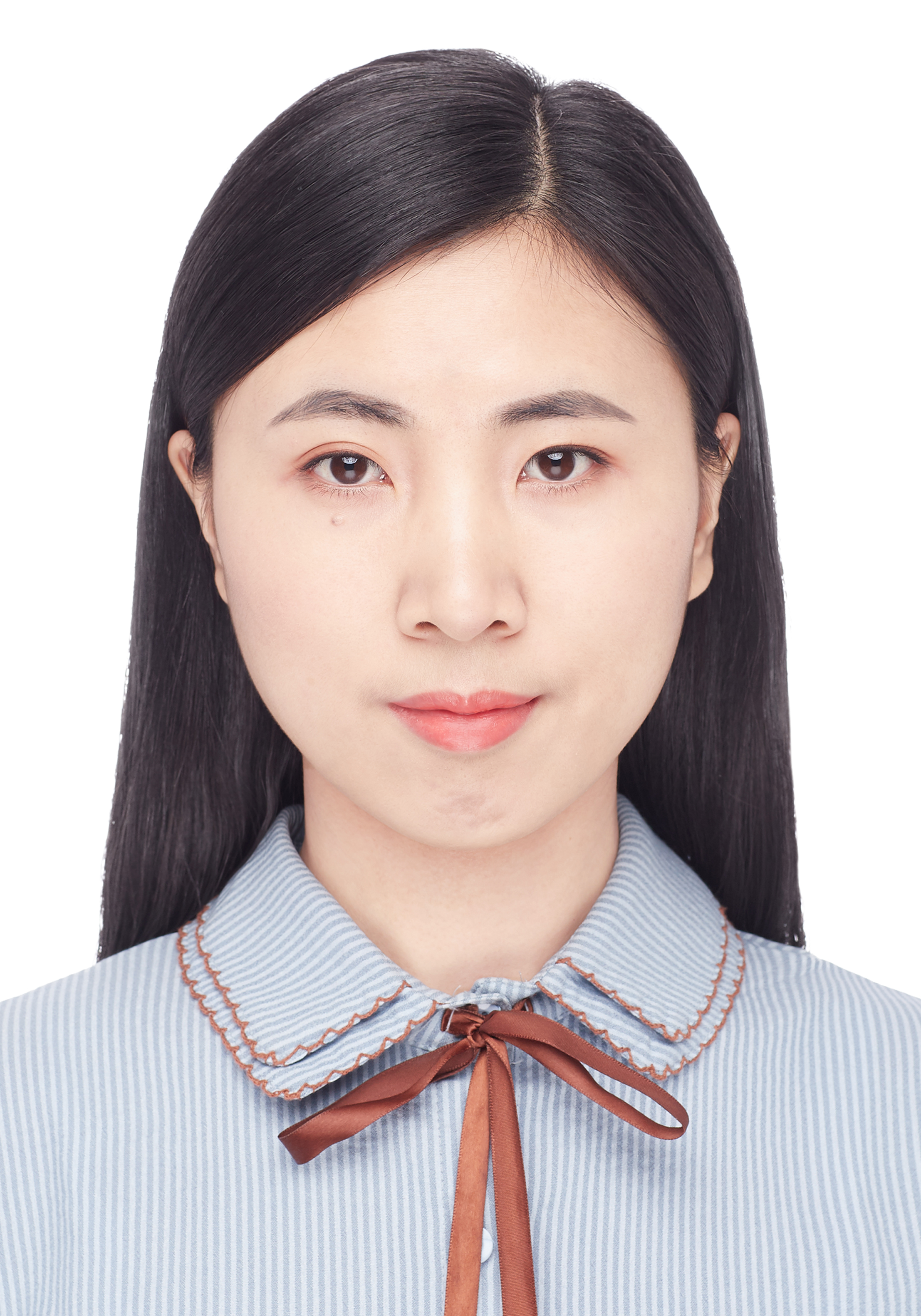}}]{Saiping Guan}
received the PhD degree in Computer Software and Theory in 2019, from the Institute of Computing Technology, Chinese Academy of Sciences, where she is currently an assistant professor. Her current research interests include knowledge graph, event logical graph, n-ary relation, etc. She has published papers in prestigious journals and conferences, including Knowledge and Information Systems, WWW, ACL, CIKM, etc. She has received the Best Student Paper Award in ICBK (2017).
\end{IEEEbiography}
\vspace{-12.5mm}
\begin{IEEEbiography}
	[{\includegraphics[width=1in,height=1.25in,clip,keepaspectratio]{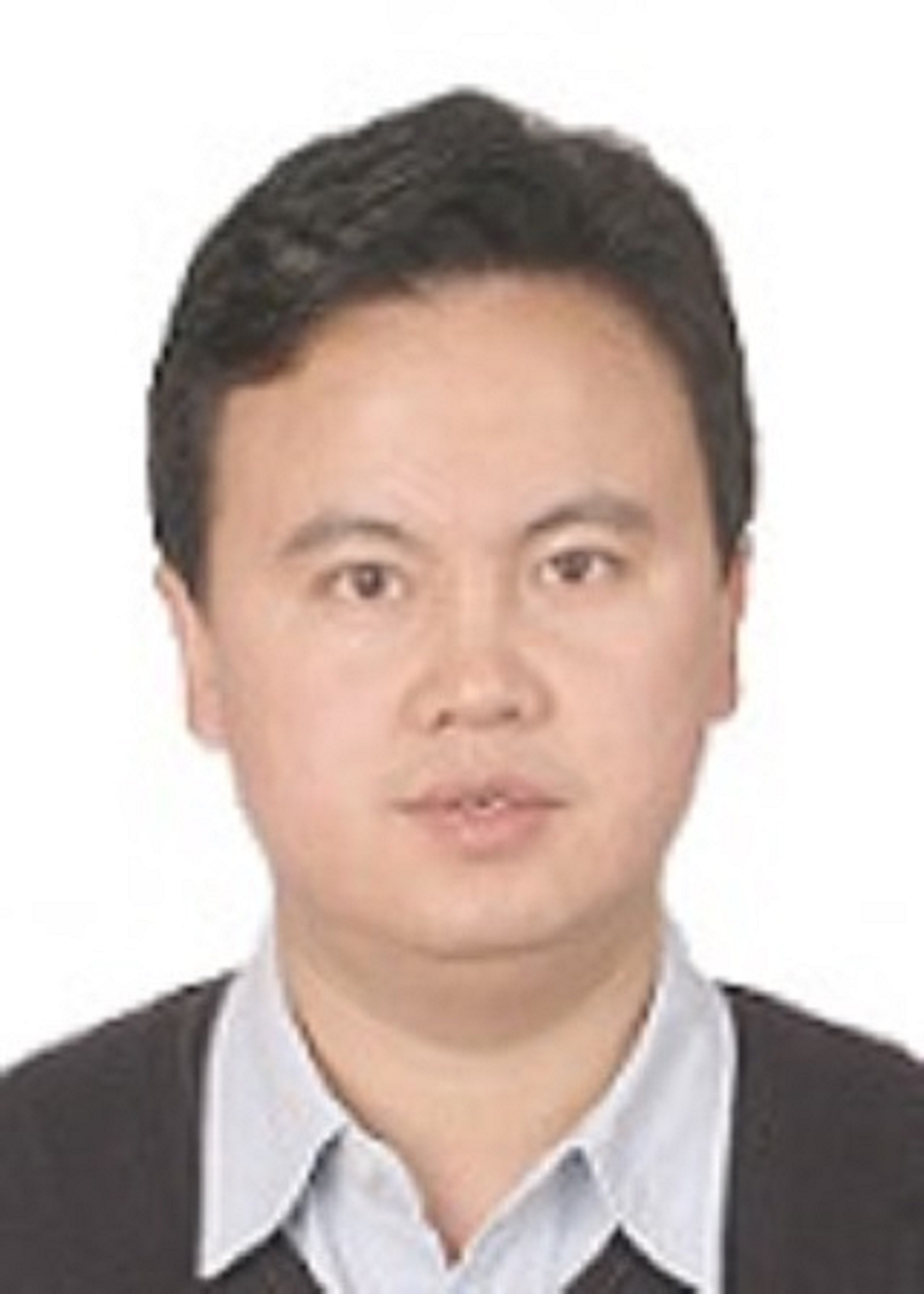}}]{Xiaolong Jin}
received the PhD degree in Computer Science from Hong Kong Baptist University, in 2005. He is currently a professor in Institute of Computing Technology, Chinese Academy of Sciences. His current research interests include knowledge graph, knowledge engineering, social computing, social networks, etc. He has published more than 200 papers in prestigious journals, including IEEE TKDE, TWC, ToC, TSMC–Part A, TPDS, IEEE/ACM TASLP, ACM TWeb, ACM TIST, etc., and in reputable international conferences, including WWW, IJCAI, AAAI, ACL, CIKM, WSDM, EMNLP, etc. He has co-authored four monographs published by Springer and Tsinghua University Press, respectively. He has received the Best (Student/Academic) Paper Awards in ICBK (2017), CIT (2015), CCF Big Data (2015), AINA (2007), and ICAMT (2003).
\end{IEEEbiography}

\vspace{-12.5mm}
\begin{IEEEbiography}
	[{\includegraphics[width=1in,height=1.25in,clip,keepaspectratio]{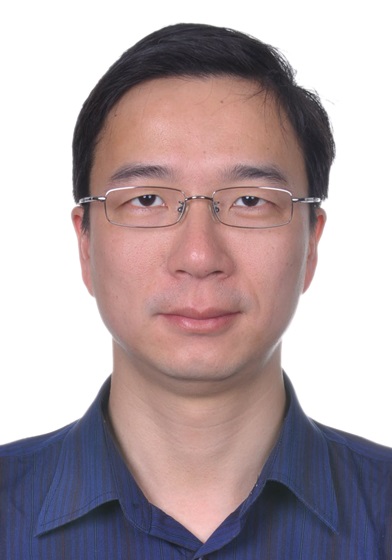}}]{Jiafeng Guo}
received the PhD degree in Computer Software and Theory in 2009, from the Institute of Computing Technology, Chinese Academy of Sciences (CAS), where he is currently a professor and the vice director of the CAS Key Laboratory of Network Data Science and Technology. He has worked on a number of topics related to Web search and data mining, including query representation and understanding, learning to rank, and text modeling. His current research is focused on representation learning and neural models for information retrieval and filtering. He has won the Best Full Paper Runner-up Award in CIKM (2017), Best Student Paper Award in SIGIR (2012), and Best Paper Award in CIKM (2011).
\end{IEEEbiography}
\vspace{-12.5mm}
\begin{IEEEbiography}
	[{\includegraphics[width=1in,height=1.25in,clip,keepaspectratio]{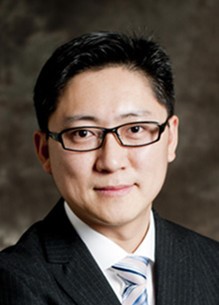}}]{Yuanzhuo Wang}
was engaged in postdoctoral research in Computer Science with the Department of Computer Science and Technology, Tsinghua University, from 2008 to 2010. He is currently a professor with the Institute of Computing Technology, Chinese Academy of Sciences. His current research interests include social computing and open knowledge network. He has authored over 160 papers. He is a Distinguished Member of the China Computer Federation. He is the local organizing chair of the conference GAME NETS 2014 and a program committee member or reviewer of NPC, ICMLC, PRDC, PDCAT, WCSE, ADHOC, SCA, CIKM, WSDM, APWEB, etc.
\end{IEEEbiography}
\vspace{-14mm}
\begin{IEEEbiography}
	[{\includegraphics[width=1in,height=1.25in,clip,keepaspectratio]{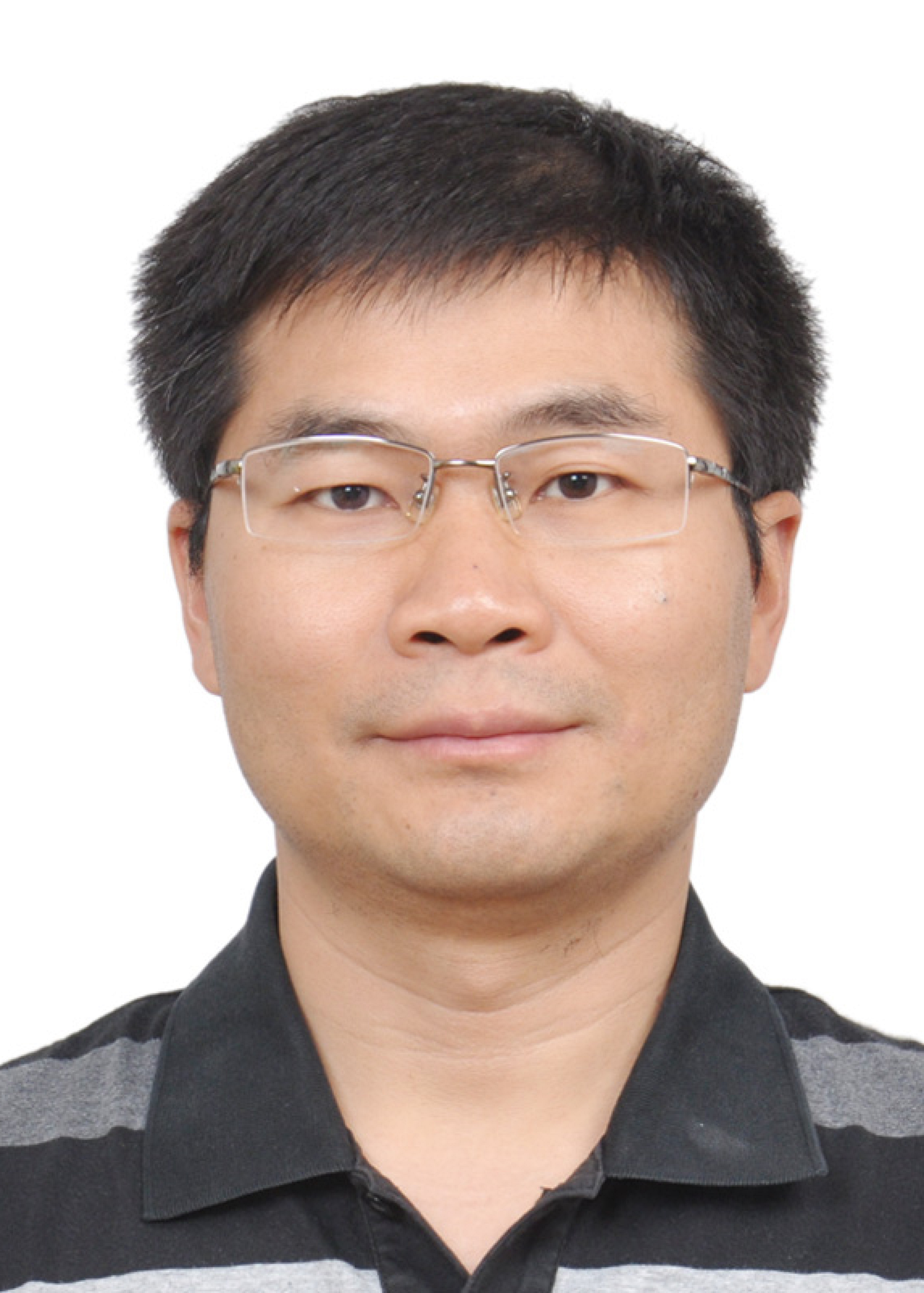}}]{Xueqi Cheng}
received the PhD degree in 2006, from the Institute of Computing Technology, Chinese Academy of Sciences (CAS), where he is currently a professor and the director of the CAS Key Laboratory of Network Data Science and Technology. His main research interests include network science, Web search and data mining, big data processing, distributed computing architecture, etc. He has published over 180 papers in prestigious journals and conferences, including ToIT, TKDE, Journal of Statistics Mechanics: Theory and Experiment, Physical Review E., SIGIR, WWW, IJCAI, AAAI, ACL, CIKM, WSDM, ICDM, etc. He has won the Best Student Paper Award in SIGIR (2012) and Best Paper Award in CIKM (2011). He is currently serving on the editorial board for the Journal of Computer Science and Technology, Journal of Computer, etc.
\end{IEEEbiography}




\end{document}